\documentclass[acmsmall, screen]{acmart}

\usepackage{booktabs}
\usepackage{amsmath,amsfonts}
\usepackage{algorithmic}
\usepackage{graphicx}
\usepackage{textcomp}
\usepackage{xcolor}
\usepackage{array}
\usepackage{balance}
\usepackage{algorithm2e}
\usepackage{subfigure}
\usepackage[english=american]{csquotes}
\usepackage{hyperref}
\usepackage{svg}

\newcommand{\approach}{{\sc HOMRS}\xspace}

\newcolumntype{M}[1]{>{\centering\arraybackslash}m{#1}}

\newcommand{\rqone}{To what extent are HMR built using HOMRS better than randomly selected MR combinations?}

\newcommand{\rqtwo}{How does HOMRS hyper-parameters influence the HMR obtained?}

\newcommand{\rqthree}{How does HOMRS compare to other generation techniques in term of coverage, number of adversarial examples generated and validity?}

\newcommand{\rqfour}{How does HOMRS compare to other generation techniques in term of time complexity?}

\AtBeginDocument{%
  \providecommand\BibTeX{{%
    \normalfont B\kern-0.5em{\scshape i\kern-0.25em b}\kern-0.8em\TeX}}}

\usepackage{ifthen}
\usepackage{xspace}

\usepackage[normalem]{ ulem }
\usepackage{soul}

\makeatletter

\newcommand{\COMMENT}[1]{}

\newcommand{\ie}{\textit{i.e.,}\xspace}
\newcommand{\eg}{\textit{e.g.,}\xspace}

\begin{document}

\title{HOMRS: High Order Metamorphic Relations Selector for Deep Neural Networks}

\author{Florian Tambon}
\email{florian-2.tambon@polymtl.ca}
\affiliation{%
  \institution{Polytechnique Montréal}
  \city{Montréal}
  \state{Québec}
  \country{Canada}
}

\author{Giuliano Antoniol}
\email{giuliano.antoniol@polymtl.ca}
\affiliation{%
  \institution{Polytechnique Montréal}
  \city{Montréal}
  \state{Québec}
  \country{Canada}
}

\author{Foutse Khomh}
\email{foutse.khomh@polymtl.ca}
\affiliation{%
  \institution{Polytechnique Montréal}
  \city{Montréal}
  \state{Québec}
  \country{Canada}
}
\renewcommand{\shortauthors}{Tambon et al.}

\begin{abstract}
Deep Neural Networks (DNN) applications are increasingly becoming a part of our everyday life, from medical applications to autonomous cars. Traditional validation of DNN relies on accuracy measures, however, the existence of adversarial examples has highlighted the limitations of these accuracy measures, raising concerns especially when DNN are integrated into safety-critical systems.

In this paper, we present HOMRS, an approach to boost metamorphic testing by automatically building a small optimized set of high order metamorphic relations from an initial set of elementary metamorphic relations. HOMRS' backbone is a multi-objective search; it exploits ideas drawn from traditional systems testing such as code coverage, test case, path diversity as well as input validation.

We applied HOMRS to MNIST/LeNet and SVHN/VGG and we report evidence that it builds a small but effective set of high-order transformations that generalize well to the input data distribution. Moreover, comparing to similar generation technique such as DeepXplore, we show that our distribution-based approach is more effective, generating valid transformations from an uncertainty quantification point of view, while requiring less computation time 
by leveraging the generalization ability of the approach.


\end{abstract}



\maketitle

\section{Introduction}\label{intro}
Deep Neural Networks (DNN), like any software program, need to be tested. This is especially necessary when DNN are used in safety-critical systems such as autonomous driving or aerospace applications, where a failure can lead to disastrous consequences. Traditional software development uses a variety of verification and validation techniques to ensure characteristics such as reliability, dependability, or safety. For example, critical avionic applications must comply with certification standards such as the DO-178C. In contrast, DNN suffer from a lack of specification and established procedure to verify them with regard to those specifications. In particular, DNN are challenging to test as they oftentimes belong to the category of programs computing an unknown answer; since the answer is unknown, they lack a mechanism to determine the correctness of the output, \ie an oracle ~\cite{DBLP:journals/cj/Weyuker82}.

Metamorphic Testing (MT)  \cite{Chen20} is a ``pseudo-oracle'' technique originally designed to alleviate the oracle problem. It has been applied both in traditional software testing \cite{Ding17} as well as in Machine Learning (ML) \cite{Xie11}; it has been adapted to test data generation \cite{Tian18}; and to detect Adversarial Examples (AE) attacks \cite{Mekala19}. Metamorphic Relations (MR) are the cornerstone of MT. A MR is a necessary property of an intended software functionality and holds true across executions. For example, a mug in a picture, remains a mug if we change the contrast, colour, or we  blur the picture. We may not know the content of a picture, but, if a DNN trained to detect mugs produces two different classifications on the original picture and on the MR transformed picture, then we know that something went wrong. MR are generally defined by researchers and through catalogs of existing MR \cite{Xie11}. Still, it is hard to know what the most effective MR will be and the notion of ``good'' relation is an ambiguous issue. Yet some rules of thumb were empirically established \cite{Segura16}, emphasizing principally on the diversity of the relations used. Recently, several approaches have been proposed to automatically identify MR, \eg{} \cite{Jie14, Chen16, Zhang19}. However, these approaches are either limited to specific types of MR (\eg{} polynomial MR \cite{Zhang19}), and/or require testers to possess solid knowledge about the problem domain (\eg{} being able to make category-choice specification for a given problem \cite{Chen16}). In DNN case, MR were generally used as a proxy to generate new adversarial cases through maximization of coverage \cite{Tian18}. As such, this approach is limited as the transformation obtained cannot necessarily generalized on any point of the data distribution. Moreover, most of those applications do not necessarily check the validity of the generated images \cite{Zhang20}, \ie{} whether the generated image belong to the input distribution. Consequently, they often 
generate images that are akin to noise for the model, and as such not relevant for testing the model. For instance, \cite{Dola21} showed by using the reconstruction probability of a Variational Autoencoder trained on the same data distribution, that previous technique such as DeepXplore \cite{Pei19} or DeepConcolic \cite{Sun19} generated many invalid images according to their metric. As the theoretical input distribution is seldom computable, validity of an image is generally checked using proxies:  human oracle, mathematical distance from an original image \cite{Guo18} or auxiliary model learning an approximation of the distribution from the training data \cite{Dola21} to name a few. However, all those methods suffer from limitations, in particular they all are limited to a point wise evaluation compared to a pre-calculated threshold, which hence cannot generalize to the validity of a transformation (MR) as a whole (\ie{} it's not clear when to consider a transformation not valid from their definition). Evaluating model's ability to understand what is presented to it, is related to the notion of \emph{uncertainty} \cite{Gal16}, as one can quantify the degree of certainty that a model has over a given input. We generalize this idea to the transformation produced over the data distribution in order to quantify the model's knowledge of the MR.  

In this paper we present HOMRS, a High Order Metamorphic Relation Selector for the generation of an optimized set of High order Metamorphic Relations (HMR) for ML/DNN testing. A High Order Metamorphic Relation (HMR) is defined via relation composition. More precisely, HOMRS represents HMR as function composition, over a set of pre-existing elementary MR. In a sense, MR are just specially case of HMR, where the composition contains only one MR. Optimized HMR could then be used to test DNN, following classical metamorphic testing, similarly to what is done for traditional software \cite{MurphyICST2008,MurphyISSTA2009}. To the best of our knowledge, no efficient approach exists for the automatic identification and selection of valid HMR for DNN.

Contrary to previous method using MR as proxy for generating new test inputs in a \enquote{depth first} manner, HOMRS thrives to generate transformation that are generalizable to the whole data distribution of the model (\ie \enquote{breadth first}). As such, any HMR set learned can be reused for any new data from the given data distribution in order to further test the model. In a way, HOMRS, is similar to techniques such as DeepEvolution \cite{BenBraiek19}, with an opposite philosophy; \ie rather than using a lot of relations to mutate a small number of test data as it was done previously, we select a small optimized set of relations that are relevant for the input distribution, in order to improve the generalization power and hence relevance. Moreover, the validity definition allows to shed some light on model's knowledge on the data distribution it learns from. This leads to HMR that are valid for the model, which make the test cases generated from the HMR relevant to probe the model for errors. Indeed, this might not be the case otherwise, as using only plain coverage metrics was shown not to correlate to test suite effectiveness \cite{Harel20}.  HOMRS works off-line on a pre-trained DNN model. A calibration set is fed to HOMRS to learn the most relevant and valid HMR via meta-heuristic multi-objective optimization. The HMR set can then be used on any unseen data from the input distribution. In order to show the effectiveness of the method, we compare HOMRS to existing generation methods; DeepXplore \cite{Pei19}, DLFuzz \cite{Guo18} and \cite{Dola21} (named \emph{DistAware} in the rest of the paper), in term of coverage, adversarial examples generated as well as validity of transformations. We selected those methods for the following reasons: DeepXplore is 
one of the first method to perform coverage guided adversarial data generation, with domain specific constraints, DLFuzz presents a fuzzing approach with the image validity being constrained by a distance metric, and DistAware follows a similar approach to DeepXplore but with the added benefit of proposing a validation method for the generated images based on a Variational Autoencoder (VAE).

We applied HOMRS to two DNNs; LeNet \cite{LeCun89} trained on the MNIST dataset \cite{LeCun98} and VGG model \cite{Simonyan15} trained on the SVHN dataset \cite{Netzer11}, as are used in compared method, and answers the following research questions:

\begin{itemize}
\item \textbf{RQ1}: {\rqone}

\item \textbf{RQ2}: {\rqtwo}

\item \textbf{RQ2}: {\rqthree}

\item \textbf{RQ2}: {\rqfour}

\end{itemize}

Our results show that HMR generated using HOMRS have a higher adversarial examples generation and coverage outperforming random sets with statistical significance. Then, we studied the effect of hyper-parameters used in our algorithm over the quality of HMR sets obtained. We also highlight that in most cases, given a similar seed, HOMRS outperform compared method (i.e., DeepXplore, DLFuzz, and DistAware). Moreover, we showed that those methods don't necessarily result in valid uncertainty profile, for the used transformation, which isn't the case for HOMRS as the optimization process take the uncertainty into account. Finally, we compared processing time between HOMRS and DLFuzz, and find that HOMRS allows for faster computation, as the computation time is seed size independent. We compare HOMRS only with DLFuzz because it is the only method that outperformed HOMRS in term of adversarial examples and neuron coverage on one of our model/dataset.

This paper makes the following contributions:
\begin{itemize}	
	\item We propose an approach for the selection of an optimized set of HMR through combination  of basic MR;
	\item We provide evidences of the effectiveness of HMR as a transformation of the data distribution with a high kill ratio while being still valid through uncertainty quantification.
	\item We study the influence of hyper-parameters of our algorithm on results obtained.
	\item We show \approach is on average more effective than similar methods, in term of coverage, adversarial examples and certainty.
	\item We highlight the advantage of generalization of our \approach comparison between our method and compared methods.
\end{itemize}
The remainder of the paper is organized as follows: \textbf{Section \ref{background}} provides background on 
the key concepts used in the approach presented in \textbf{Section \ref{approach}}. 
\textbf{Section \ref{experiments}} reports about the experiments performed to evaluate our proposed approach and \textbf{Section \ref{results}} discusses the results of those evaluations. \textbf{Section \ref{discussion}} discusses our results and approach while \textbf{Section \ref{threats}} presents the threats to the validity of our study. \textbf{Section \ref{related}} gives a short overview of related works and \textbf{Section \ref{conclusion}} concludes the paper.


\section{Background}\label{background}
This section is intended to provide a brief description of concepts deemed useful to understand our proposed approach HOMRS.

\subsection{Deep Neural Networks}

DNN are composed of multiple layers of neurons that are tasked with elaborating information from provided  inputs. In the basic case of fully-connected DNN, each neuron can be viewed as a function $f$, whose inputs are either the input data or the output of other neurons, typically a previous layer. Each neuron has a set of weights $\textbf{w} = {w_1, w_2, ..., w_n}$ where $n$ is the number of inputs and a bias $b$. If  $\textbf{x}$  is the neuron input vector, the output of the neuron is computed as $f(\textbf{x}) = \sum_{i=0}^n w_ix_i + b$. 
Neurons outputs are transformed via a non linear \textit{activation function} ``complexifying" the behaviour of the network. Overall, the output of a neuron is computed as $g(f(\textbf{x}))$, where $g$ is an activation function such as the ReLU function \cite{Glorot11}. The input vector is propagated throughout the network until the output where the loss function $L$ is used to compute the distance to the desired target. From there, a back-propagation algorithm can be used \cite{Rumelhart86} to update the weights. To train DNNs, in traditional settings, the initial dataset is generally split into a train and test set (and possibly a validation set which is used to test for overfitting or hyper-parameters tuning). The model is trained on the training set and tested on the test set to evaluate its performance on unseen data (generalizability).  

\subsection{Adversarial Examples and Out-Of-Distribution Examples}

Given a DNN $f$ and an image $\textbf{x}$ with its ground truth class $\textbf{y}$, an \textit{adversarial example} is defined as an image $\textbf{x'}$ such as $\lVert \textbf{x'} - \textbf{x} \rVert < \delta \implies f(\textbf{x'}) \neq \textbf{y}$, where $\delta$ is relatively small and $\lVert . \rVert$ is a given distance metric \cite{Yuan19}. It basically includes all images close to $\textbf{x}$ for the defined distance but yielding a different prediction through the DNN. We distinguish them from \textit{out-of-distribution} (OOD) examples which are any anomalies too different from the dataset (``in") distribution the model was trained on.

\subsection{Uncertainty estimation}

Uncertainty estimation are used as a measure of the confidence of the prediction the model is making. As the output of the network in classification task settings is a softmax vector, it could be tempting to assimilate the vector as probability one and take the highest value as a form of certainty in the model's prediction. However, using plain softmax vector for this purpose was shown to be misleading \cite{Pearce21}. In practice, DNN's uncertainty can be estimated using Monte-Carlo sampling with Dropout at inference time \cite{Gal16}, \ie MCDropout. In practice, $N$ forward passes through the network are performed with Dropout (\ie randomly deactivating neurons given a probability $p$) turned on and the results are averaged. This approach is in a sense similar to model ensemble and was proven to approximate Bayesian Neural Network, and thus model's uncertainty. This value can in turn be used for instance for anomaly/OOD detection \cite{Liang20}, where a threshold of certainty can be defined to separate data from the actual input distribution and anomaly.

\subsection{Multi-objective optimization and NSGA-II}

Multi-objective optimization aims to tackle problems with conflicting objectives. In our context, we would like to keep the set of HMR small,  but we want it to be very effective in transforming images while also  exercising as much as possible the DNN logic. Multi-objective problems are generally defined as:
\begin{align*}
	\text{min/max} \quad f_m(\textbf{x}), i = 1,...,M \\
	\text{Subject to}: g_j(\textbf{x}) \geq 0, j = 1,...,J \\
	h_k(\textbf{x}) = 0, k = 1,..., K  \\
	x_i^{(L)} \leq x_i \leq  x_i^{(U)}, i = 1, ..., n
\end{align*}

where $\textbf{x} = {x_1, ..., x_n}$ represents a solution, $f_m$ the $M$ objectives, $g_j$ and $h_k$ the constraints of the problem and $x_i^L$ and $x_i^U$ the lower and upper boundaries of the $i$-th component of $\textbf{x}$. 

HOMRS applies NSGA-II (Nondominated Sorting Genetic Algorithm II)  to  build  a Pareto-front, which is the set of non-dominated solutions, where each point
represent a solution; hence a HMR set. More formally, a solution $x$ is \emph{non-dominated}, if $\forall x' \in Solutions, \exists i \in [1, M], f_i(x) > f_i(x')$ (resp $<$ if objective is minimized), \ie the solution $x$ is better than any other solution on at least one criteria.
NSGA-II is based on the Genetic Algorithm but applied to a multi-objectives problem. It follows the same principle of \textit{Evaluation}, \textit{Selection}, \textit{Crossover}, and \textit{Mutation} of the base algorithm, but uses: (1) an \textit{elitist} principle, which means that the best solutions of one generation can be carried to the next one, (2) sorting non-dominated solutions through Pareto-front optimization, (3) the \textit{crowded-distance} that maintains population diversity by spreading uniformly the Pareto-front.

\subsection{Metamorphic Testing \& Relations}\label{meta}

Let's consider the basic $sine$ example from \cite{Chen20}; where 
$P$ is a program implementing the $sine$ function. The goal is to test $P$ to see if it matches the mathematical function. Now it might not always be possible to know the exact value of the function for all possible angle values; i.e., the correct value of $sin(x)$ for every given $x$ (the ``oracle'' problem). However, it is known that $sin(x + \pi) = - sin(x)$. The relation $x + \pi$ is what is called a Metamorphic Relation. Leveraging this relation, one could compute 
for a given \textit{source} test input data $x$, a \textit{follow-up} test data $x + \pi$, and 
verify that $P(x + \pi) = - P(x)$ to test our implementation, without the need for the ground-truth. 
The identification of these MR is problem and data dependant. When images are the data and object detection is the problem, MR should preserve the nature of objects in the images. Simple images based MR include translation, rotation, shear, scaling, blurring, and contrast. 

HMR can be obtained through ``chain" of simple relations \cite{Wu05}. Let $f_1, ..., f_n$ be $n$ simple MR; a chained relation is defined as the composition of any number of those $g : f_1 \circ ...\circ f_n$. Note that obviously, the functions need to be composable, that is for $f_j \circ f_i$ we need to have that $\forall x \in D_i$, $f_i(x) \in D_j$ where $D_i, D_j$ are the domain of definition of $f_i, f_j$. In our case, since we are dealing with image based transformations, the domain of definition is the same for all transformations. However, contrary to classical chained relations \cite{Wu05}, where $f_i$ are applied to incrementally generate test data and incrementally test the program, we don't aim for the single, known, $f_i$, rather for $g$ the final composition; \ie  intermediate images are not of interest here.

\subsection{DNN Test Generation methods}

In this work, we will compare our method to three test generation methods for DNN. DeepXplore \cite{Pei19} is a white-box differential testing approach, using multiple DNN, which aim to jointly optimize a problem of maximizing neuron coverage and the prediction differences in between the DNN through gradient ascent with predefined transformations. On the other hand, DLFuzz \cite{Guo18} uses fuzzing technique to generate new images, through neuron coverage based adversarial generation. DLFuzz limits the perturbations induced on the generated images (similarly to what one could expect of some adversarial examples) in order to preserve the validity of the images. Finally, \cite{Dola21} uses an approach analog to DeepXplore with the extra step of a VAE trained on the same distribution as the models which become a part of the joint optimization process. This allows them to check for the validity of the generated test cases w.r.t to the training distribution the VAE learned, which differ from the two other approaches. We selected those methods for the following reasons: DeepXplore is one of the first method to perform coverage guided adversarial data generation, with domain specific constraints, DLFuzz presents a fuzzing approach with the image validity being constrained by a distance metric, and DistAware follows a similar approach to DeepXplore but with the added benefit of proposing a validation method for the generated images based on a Variational Autoencoder (VAE).


\section{Approach}\label{approach}
\begin{figure}
    \centering
    \includegraphics[width=1.2\textwidth]{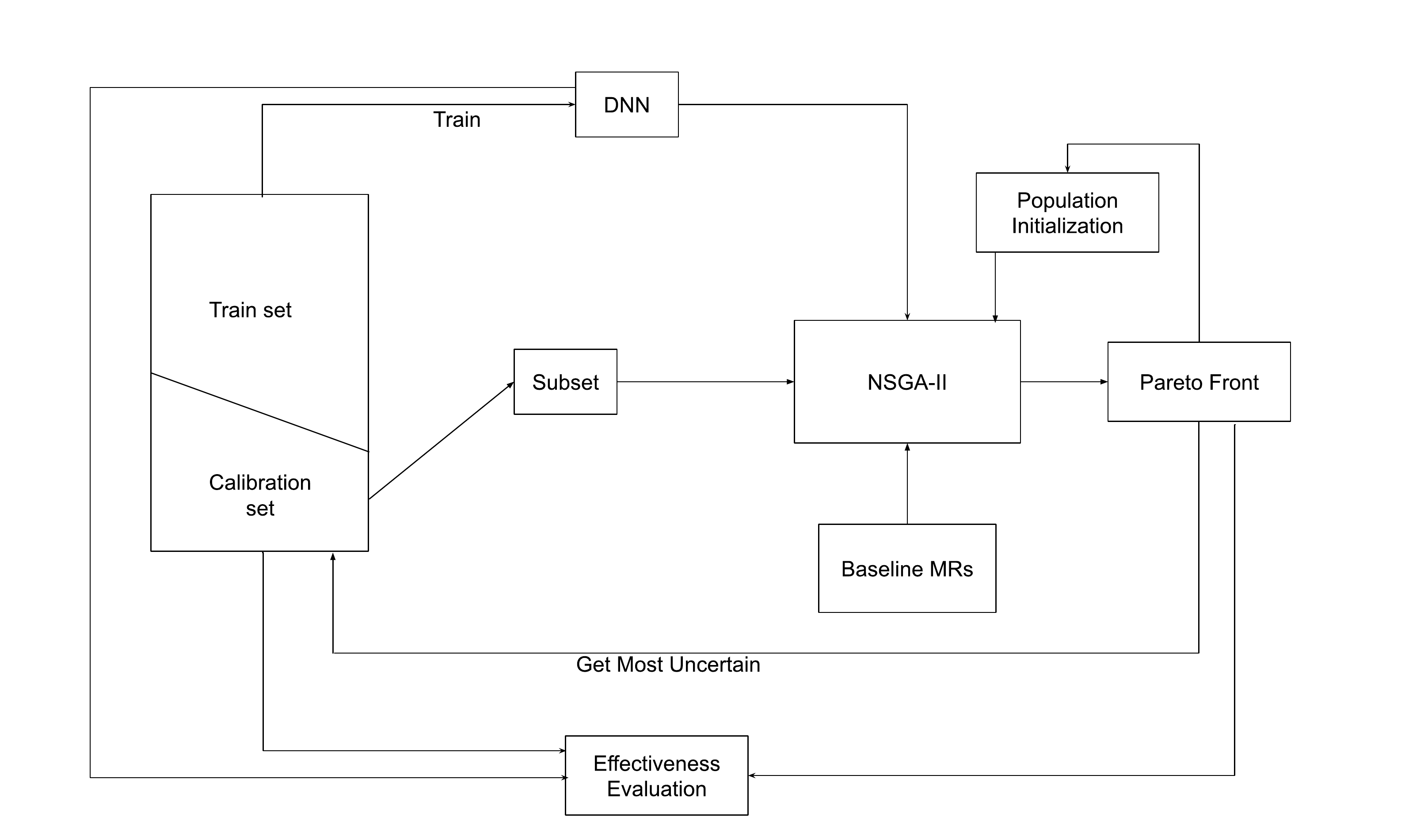}
    \caption{HOMRS high level view}
    \label{diagram}
\end{figure}

As shown in Fig. \ref{diagram}, the corner stone of HOMRS is a multi-objective optimization. 
In the following, we present a detailed description of HOMRS  
in the context of DNN, image classification, and NSGA-II, with a brief pseudo-algorithm presented in Algorithm \ref{algo}.


\IncMargin{1em}
\begin{algorithm}[h]
 \SetKwData{Left}{left}\SetKwData{This}{this}\SetKwData{Up}{up}
 \SetKwFunction{ComputeUncertaintyThreshold}{Compute Uncertainty Threshold}
 \SetKwFunction{Opt}{Optimize}
 \SetKwFunction{GetMostUncertain}{GetMostUncertain}
 \SetKwInOut{Input}{input}\SetKwInOut{Output}{output}
 \Input{DNN to test $\mathcal{D}$, a calibration set $\mathcal{S}$, a set of base MR $\mathcal{M}$ and their parameters boundary, $n$ the number of reruns}
 \Output{A set of optimized HMR $\mathcal{M}^*$}
 threshold $\leftarrow$ \ComputeUncertaintyThreshold($\mathcal{D}$, $\mathcal{S}$)\;
 $s$ $\leftarrow$ Random($\mathcal{S}$)\;
 $i \leftarrow 0$\;
 \While{True}{
  $\mathcal{M}^*$ $\leftarrow$ \Opt($\mathcal{D}$, $s$, threshold)\;
  \If{$i = n$}{
   break\;
   }
  $s$ $\leftarrow$ \GetMostUncertain($\mathcal{D}$, $\mathcal{S}$)\;
  i++\;
  
 }
 \caption{Algorithm to select the HMR}\label{algo}
\end{algorithm}\DecMargin{1em}

HOMRS takes as input a calibration dataset, a set of MR relations altogether with relations bounded parameters ranges and a trained DNN architecture. The calibration dataset is akin to a validation set, and is used by HOMRS to optimize the HMR. HOMRS represents an \textit{individual}, a  \textit{solution}, as a tree containing one or more HMR (\textit{genes}). A HMR is a path from the root of the tree to a leaf. As computing on the whole dataset is computationally expensive, HOMRS first samples randomly a subset of size $S$, from the calibration dataset. On this sub-set the meta-heuristic algorithm returns the sets of HMR, \ie the $g$ chained relations built form the pool of input relations, on the Pareto front. Those obtained HMR are then used to compute the plain uncertainty over the calibration dataset transformed with the HMR. This allows to generate a new subset, where a part of the subset is composed of the $p\%$ most uncertain data for those HMR, $p$ being an hyper-parameter, the rest of the subset being selected at random. This subset is then reused as input for the meta-heuristic algorithm, with the obtained HMR being used as the initial population of this new iteration. The process is repeated for a given number of steps $N$.

Note that complexity increases with the number of HMR $K$, HMR maximal tree depth $D$, and the number of basic types of MR $n$. From the step 2, subsets are composed partly of the most uncertain data with regards to the HMR. The assumption is that, since we limit ourselves to small subset and in order to get valid transformations over the whole data distribution, we need to overestimate the potential uncertainty of the HMR by looking mostly at data yielding high uncertainty.

HOMRS runs a multi-objective optimization on the subset guided by the fitness functions. Sub-set data is input to the DNN. The DNN execution is monitored to extract neural patterns. The extracted information and the output of the network guide the search. The search output is a set of Pareto fronts. In order to increase the relevance of the solutions, HOMRS uses a special kind of elitism. The algorithm is restarted $N$ times, each time with a subset of the most uncertain data with regards to the HMR of the current Pareto front. At step $i$, the Pareto front is stored but also used to initialize the population at step $i+1$. The idea is that since we obtain a certain front for a subset, it is likely that at least some of those solutions can also be solutions for another subset (or at least be close to the solutions), even if the new subset contains uncertain data. At the end, a final Pareto front is returned with HMR sets. The set can then be evaluated on the whole calibration set. Finally, the user can then choose which set is more fitting to its requirements (highest coverage, highest kill ratio, balanced approach...).

\subsection{Solution Representation and Genetic Operators}\label{nsga_impl}

\begin{figure}[ht]
\centerline{\includegraphics[width=0.50\textwidth]{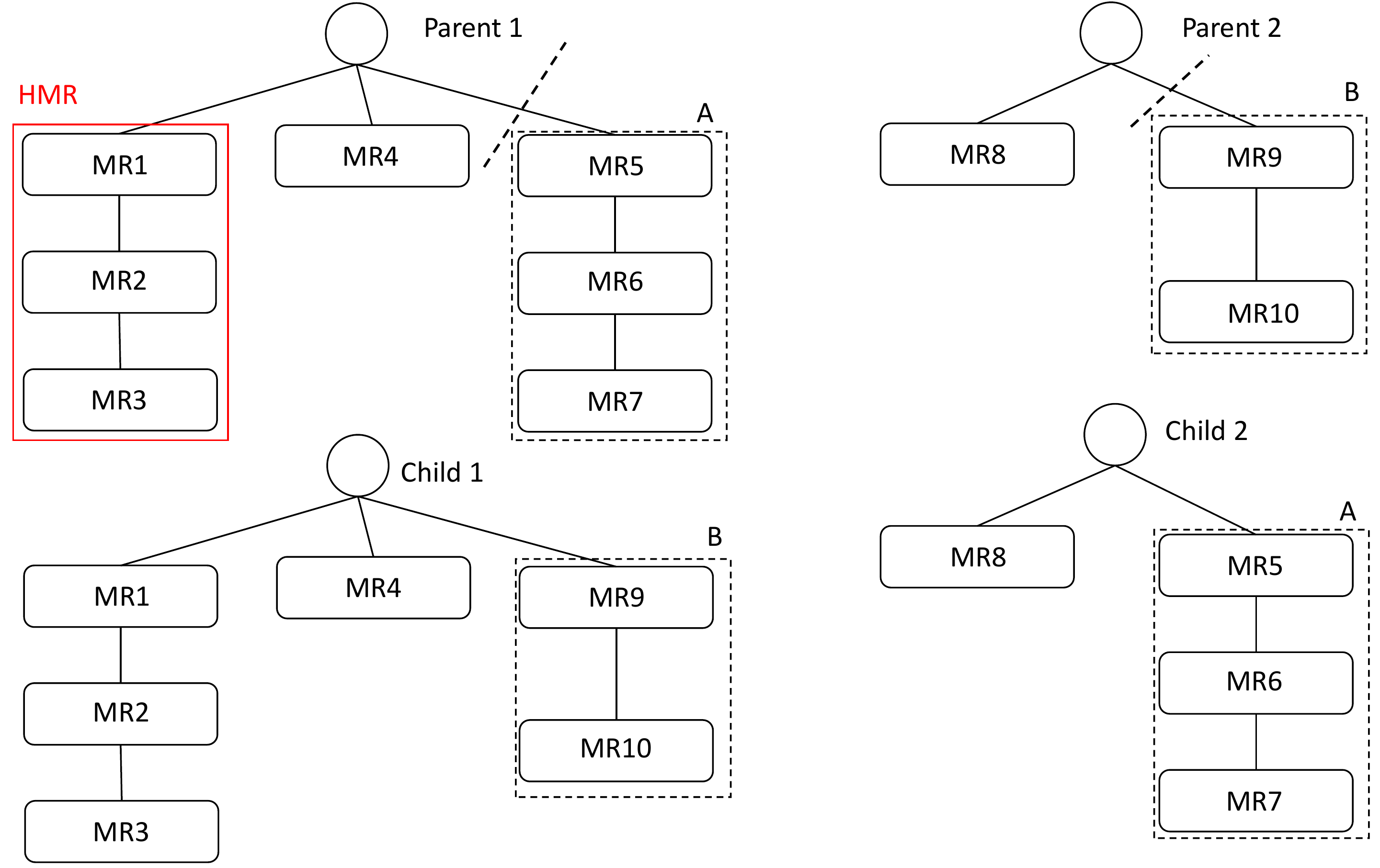}}
\caption{Single-point crossover operation with examples of individuals. Each individual is composed of multiple HMR that are the branches of the tree. For instance, the individual represented by ``Parent 1" has an HMR [MR1, MR2, MR3] (red), where each MR is a basic metamorphic relation.}
\label{chromosome}
\end{figure}

HOMRS individuals are set of chains of MR, they are rooted trees. A representation of individuals plus the crossover operator are  shown in Fig. \ref{chromosome}. HOMRS  tree root has a variable number, at most $K$, of children branching from it. A root-to-leaf path is  a HMR.  $K$ is our $budget$ number of HMR in an individual. 

Crossover operates on gene paths, swapping HMR between individuals.  Mutation operator is subdivided into three sub-operators. Mutation  can either change the value of the parameters (but keep the same relation), nullify the relation (i.e., considering that it is not activated) or reinitialize the relation (new parameters and potentially new relation).

HOMRS restricts the number  of relations present in a solution, thus limiting the HMR execution time. Note that even though we have a maximum budget, we do allow (and encourage) the number of relations used in a solution to be as small as possible while maximizing the three objectives.

\subsection{Multi-Objective Fitness Function}\label{criteria}


To the best of our knowledge there is no clear and formal definition of the notion of MR quality for DNN. Intuitively a good HMR should fulfill the pseudo-oracle task for a DNN. This in turns means that the output of DNN on the original datum, image, and the transformed datum will be different. To assess the quality of HMR in the context of DNN, we adopt criteria proposed by Segura et al. \cite{Segura16} and adapt concepts of traditional software testing such as code coverage, path coverage, and test case diversity. 

The first objective function is inspired by code coverage; the more code is tested, the more likely defect will be exposed. Any coverage criteria adapted to DNN can be used, but we will consider only two criteria that we used in our experiment, that is Neuron Coverage (NC)\cite{Pei19} and Distance Surprise Adequacy (DSA)\cite{Jinhan19}. As NC is mainly used in comparison of Section \ref{experiments}, we present here the definition used: for a test (sub)set $\mathcal{T}$, a number of activated neurons $N_{act}$, and a total number of neurons $N_{tot}$, NC is defined as: 

\begin{equation}
	NC(\mathcal{T}) = \frac{N_{act}(\mathcal{T})}{N_{tot}}\nonumber
\end{equation}
 
The second objective function has the goal to mimic  path coverage and test case diversity. We want to maximize the diversity among HMR while ensuring that different DNN execution paths are exercised.
HOMRS adapts the idea of \textit{path coverage} to  DNN in the following way. First we define a measure of \textit{neuron similarity}. Let $t_0$ be a non transformed original test case, and let $t_k$ with $k = 1, ..., K$, where $K$ is the number of HMR to be applied or \textit{follow-up} test cases. In a nutshell, each $t_k$ is the result of the application to $t_0$ of a candidate HMR of an individual of the population. Let $\textbf{t} = [t_0, t_1, ..., t_{K}]$, we define the neuron similarity of $\textbf{t}$:

\begin{equation}
	Nsim(\textbf{t}) = \frac{1}{N_{tot}\binom{K+1}{2}}\sum_{\substack{t_i \in \textbf{t}\\ i \neq j}}\sum_{t_j} \lVert a(t_i) - a(t_j) \rVert \nonumber
\end{equation}

where $\lVert . \rVert$ is a distance (Hammin, L1, ...) defined depending on the coverage criteria used and $a(t_i)$ represents the coverage path of the test $t_i$ through the DNN. The similarity is defined with regards to the coverage criterion used. If NC is used for coverage, it captures the state of activation of each neuron when a test $t_i$ is passed through the DNN. If it's DSA, it quantifies the surprise value of each neuron when a test $t_i$ is passed through the DNN. This measure quantifies how similar are HMR (and the original test) to each other. We divide it by the number of neurons and combinations in order to obtain a value between 0 and 1. The similarity over the whole test (sub)set $\mathcal{T}$ is then:

\begin{equation}
	NSim(\mathcal{T}) = \frac{1}{|\mathcal{T}|}\sum_{\textbf{t}_i \in \mathcal{T}}Nsim(\textbf{t}_i) \nonumber
\end{equation}
    
where $\textbf{t}_i$ represents the original/follow-up tests for each test of the test subset inputted to the DNN. This criterion allows for a better spread of the tests path coverage, a better diversity of the paths triggered by each relation. Moreover, it also fills the role of a normal \enquote{size} criterion since it penalizes set of relations where the number of relations is too high, as the more relations there is, the more likely there is to be some similarity.

The last objective function is inspired by metamorphic testing and measures the error detection effectiveness of the HMR. HOMRS checks if a test returned a different output through the DNN compared to the output of one of the follow-up tests. The \textit{kill ratio} over the test subset $\mathcal{T}$ is computed as:

\begin{equation*}
	KR(\mathcal{T}) = \frac{1}{|\mathcal{T}|}\sum_{\textbf{t}_i \in \mathcal{T}} \mathcal{B}(\textbf{t}_i)
\end{equation*}

where, $\mathcal{B}$ is a boolean function returning $1$ if there exists a relation for which the MT fails, $0$ otherwise.
We average over the whole (sub)set size to have numbers between zero and one. The \textit{kill ratio} measures how many \textit{unique} tests fail: one or more relations returning an error for a given source test will still count as one, as just one relation is needed to detect an error.

Overall, HOMRS uses three objective functions 1) to maximize coverage, 2) increase diversity of tests (by minimizing similarity), and 3) maximize the number of unique errors found by HMR sets. 

\subsection{Uncertainty constraint}\label{unc_const}

As mentioned in the introduction, some techniques proposed a mechanism for checking the validity of generated images. However, all these mechanisms present some short-comings. Human oracle, while being the standard in many domains, can be outwitted on some tasks and cannot be automated or scaled. Distance ($L2, L1...$) based validity is detached from model's knowledge and needs to be considered based on task, data type, and transformations at hand. Finally, using auxiliary models such as VAE in DistAware consists in relying on a third party model to validate outputs, with approximations resulting from the reconstruction. In our context of MT, the biggest downside of these methods comes from the fact that they cannot be applied straightforwardly to assess the validity of a \emph{transformation}, as they are all point-wise evaluations.

\begin{figure}[ht]
  
\centerline{\includegraphics[width=0.60\textwidth]{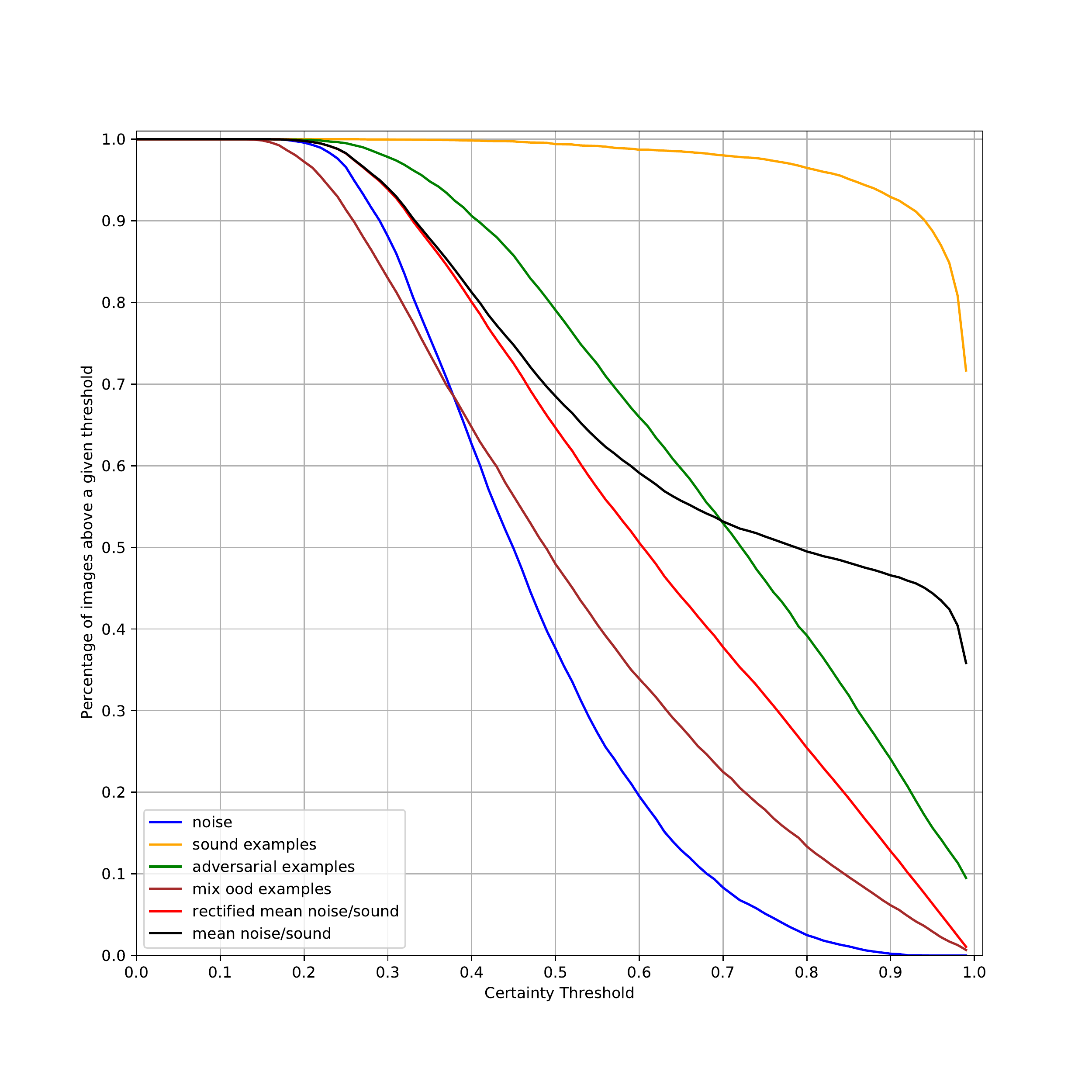}}
\caption{Uncertainty profiles plot (MNIST/LeNet). Black line is the average of noise and sound data uncertainty. Red line is a rectified average. Red dots shows the two calculated points returned by function \emph{Uncertainty Threshold}. Green line is the profile of adversarial examples generated from calibration set and FGSM techniques. Brown line is the profile of OOD data gathered from datasets different from the input distribution. Note, we define \emph{certainty} as $1 - \emph{uncertainty}$}
\label{unc_prof}
\end{figure}

In Section \ref{background}, we presented uncertainty as proxy of the model knowledge, that is, its level of confidence in its prediction? and thus the understanding he has of an input. However, uncertainty cannot be used as it is done in traditional settings, i.e., by fixing a threshold and cutting input which lead to a higher value, since doing this would lead to the same problem faced by other validation methods. Specifically, all methods we compare our approach to use point-wise estimation with, as previously mentioned, a cut threshold. Thus, if we want to quantify validity of a whole \emph{transformation} using their approach, it's not clear how one could do, \ie which amount of data of the transformed distribution would be required to be above the calculated threshold for us to consider the transformation valid. As our aim is to have generalizable valid transformation, we will instead rely on \emph{uncertainty profiles} to obtain a lower bound of the uncertainty profile that a HMR set can have. This will in turn act as a constraint for our algorithm. Note that in the following we define \emph{certainty} as $1 - \emph{uncertainty}$.

We define an \emph{uncertainty profile} as a function of the percentage of images of a given input distribution that are above a certain certainty threshold $t$ such as on Figure \ref{unc_prof}. As we want to obtain a lower bound of distribution validity, we will use two profiles that we can infer from the task of the model; the calibration dataset and noise dataset. The first one is composed of data that we know are valid (since the data are essentially similar to the distribution the model was trained on), while the second one should not be valid, as the data are too noisy thus close to meaningless. Those two profiles are the orange and blue curves on the Figure \ref{unc_prof}. We use them in order to compute a mean of their profiles (the black line), which roughly represents a lower bound of validity for any distribution w.r.t our input distribution. However, as one can see, this leads to a relatively high percentage of images for which the model has a quasi certainty, which is not desirable for a lower bound, as the generated transformations should constitute the least valid among valid distributions possible. Hence, in order to rectify the curve, we use a smoothing average function, defining the certainty function $\mathcal{C}$:
\begin{equation*}
   \mathcal{C}: t \longrightarrow \frac{1}{2}((1 - t^3) \times u + (1 + t^3) \times l)
\end{equation*}
where $t$ is the current certainty threshold, $u$ the number of normal images (non-transformed) from the dataset above the threshold $t$ and $l$ the number of noisy images above the same threshold. This leads to the more desirable red curve. The polynomial expression's degree was chosen empirically based on trial and errors, in order to smoothen the empirical mean to obtain the desired property.

If this lower bound is empirically obtained, we provide additional empirical evidence to show its relevance. We plotted two additional profiles from two other distributions: the brown one is the uncertainty profile of OOD data, created from a mix of multiple dataset different from \enquote{in}-distribution one (see Section \ref{experiments}), while the green one is the profile of Adversarial data generated using the calibration set and the FGSM \cite{Goodfellow14} method. That is, we have a profile of one distribution of invalid data (OOD) which as such should result in a lower certainty profile, while the second distribution is based on valid data that are specifically crafted to confuse the model while being part of its distribution. As such, we see our lower bound profile clearly separating the two distributions, OOD on one part and Adversarial Examples on the other. Hence, similarly to Adversarial Examples, our HMR transformations should result in a certainty profile above this line.

Hence, this empirical threshold will serve as a hard constraint to be used in our algorithm in order to validate whether a transformation is valid or not, that is if its uncertainty profile is above the empirical threshold. 


\section{Evaluation}\label{experiments}
We assess the effectiveness of HOMRS by answering the 
following Research Questions (RQs): 

\begin{description}
\item[\textbf{RQ1.}] \rqone
\item[\textbf{RQ2.}] \rqtwo
\item[\textbf{RQ3.}] \rqthree
\item[\textbf{RQ4.}] \rqfour

\end{description}

\subsection{Procedure}\label{procedure}
Each  RQs required a specific setup detailed in to following.

\begin{itemize}
\item To study RQ1, we used a LeNet5 (resp. VGG-19) model trained over the MNIST (resp. SVHN) dataset. Both models were modified to include the Dropout layers following a procedure similar to \cite{Gal17}. We apply \approach following the block diagram  of  Fig \ref{diagram}. We train our model using SGD optimizer with default parameters. The achieved test set accuracy was 98.97\% for LeNet and 96.08\% for VGG. The used MR are those listed in Table \ref{transf} which are MR previously used in \cite{Tian18}.

The range parameters of MR were selected to limit the search space, but can be extended to any range in theory as the uncertainty threshold would reject transformation with parameters that modify images too much. We ran $5$ steps with each having a size of 10\% of the calibration dataset size, in order to have a small enough subset for computation but with a good enough representativity. We fix in all cases a budget of five (so all the individuals are of size five). We set HMR chains depth to three. The budget was chosen based on empirical studies \cite{Liu14} suggesting that three to six diverse MR  are enough to reveal a large  number of faults. Depth was chosen in order to avoid increasing the time complexity too much. To quantify the effectiveness of \approach  against randomly generated HMR sets, we used \approach $10$ times in order to generate $10$ Pareto fronts from which we choose the most balance set and we randomly sampled $30$ HMR sets to compare to.

For evaluating Neuron Coverage (NC), we fixed the activation threshold to $0.25$ as is used in the methods we compare ours to. For Distance Surprise Adequacy (DSA), we used in both case the same parameters that were used in the paper introducing the criteria \cite{Jinhan19}, that is $1,000$ buckets and $2.0$ as the upper bounds.

We then applied the MC Dropout procedure on multiple datasets in order to obtain uncertainty profiles as detailed in Section \ref{unc_const}. We used for all models the calibration set to compute the profile of normal data and noisy data and we used test data to compute for the chosen HMR transformed data distributions. We also used test data adversarially modified with FGSM ($\epsilon = 0.2$ for MNIST, $\epsilon = 0.05$ for SVHN) as well as OOD datasets. For OOD, all models used one made of mixed dataset (FashionMNIST\cite{Xiao17}, NotMNIST\cite{Bulatov11}, Omniglot\cite{Lake15} and CIFAR-10\cite{Krizhevsky}) and one specific: for LeNet/MNIST we used NotMNIST and for VGG/SVHN we used CIFAR-10, as they are quite similar to the input dataset. 

\item RQ2 : To study RQ2, we used MNIST/LeNet with NC criteria, in order to have an acceptable computing time due to the multiple runs necessary. The goal was to study how the hyper-parameters of our algorithm influence the HMR obtained. Said parameters encompass the number of subset processed by the NSGA-II $a$, the number of evaluations of the NSGA-II $b$ and the percentage of most uncertain data part of each subset $c$. Grid of parameters used can be found in Table \ref{param}. Each individual parameters settings was repeated 3 times in order to tackle randomness.

\begin{table}[!t] 
\caption{Hyper-parameters used in RQ2 study}
\renewcommand{\arraystretch}{1.2}
\centering
\begin{tabular}{c|c|}
\hline
$a$ & $3, 5, 10$\\
\hline
$b$ & $100, 200, 400$\\
\hline
$c$ & $0.01, 0.02, 0.04$\\
\hline
\end{tabular}
\label{param}
\end{table}

\item RQ3: we will study \approach effectiveness against other generation techniques: DLFuzz, DeepXplore and DistAware. In all case, we used default parameters of each method as well as the implementation provided. All algorithms used the same randomly sampled seed of size $500$. All algorithm ran until all inputs were processed according to the algorithm. As DSA is not evaluated in those methods, we will only compare using NC with the same threshold of $0.25$ that was used in the previous paper. We evaluated each method using 
the Neuron Coverage, Adversarial Examples generated, and uncertainty profile following the same MC Dropout method as HOMRS. 

\item RQ4 : we will compare time complexity between our method and DLFuzz. We selected DLFuzz because it is the method with the closest score (coverage and \#Adv) to ours in RQ3, even beating our method in \#Adv for SVHN/VGG on the seed of size $500$. We reuse the 10 runs we did for our method for SVHN/VGG in previous questions and average the time obtained for each of the independent runs. We independently ran DLFuzz 4 times as the algorithm requires more time for each run. Both method were ran on the same configuration (AlmaLinux release 8.4, RAM: 32GB, CPU: AMD Ryzen 7 3800X 8-Core Processor, GPU: NVIDIA GeForce RTX 3080). As DLFuzz implementation used an older version of Tensorflow compared to the version we are using (Tensorflow 2.5), we updated the code of DLFuzz to allow it to run with the new version. We did a sanity check with the results obtained to ensure that updated version didn't alter the working of DLFuzz algorithm. Note that we measured in both cases, the algorithm's processing time. Note that \approach processing time computed doesn't take into account the time it takes once HMR are selected to generate on a given seed the new image, only the processing time of the algorithm to obtained the optimized HMR. However, in practice, this time is negligible w.r.t to the execution time of the algorithm (a few seconds on several thousands of images) when using GPU/batch generation.


\end{itemize}

\subsection{Variable Selection}
To answer the research questions we measured different variables. More precisely:
\begin{itemize}
  
\item To answer RQ1 we measured the coverage (NC/DSA), similarity and kill ratio; we did  this for \approach as well as the randomly generated HMR. We also used statistical tests (Mann-Whitney and Cliff's delta) using those results. We then used chosen HMR sets derived by \approach to compute uncertainty profile over each datasets distribution in order to compare them.
  
\item To answer? RQ2, we measured the same criteria as RQ1 (although only NC) in order to be able to verify how the parameters would influence criteria.
 
\item To answer RQ3, we compare our method to the others previously listed in term of coverage/adversarial examples and uncertainty.

\item To answer RQ4, we did a sanity check by verifying that results (coverage and number of images generated) obtained between the original DLFuzz implemetation and the updated version were similar. We then compared processing time of both DLFuzz updated version and \approach.
  
\end{itemize}

%

\subsection{Instrumentation and NSGA-II parameters}
To carry out the experiment, we implemented a \approach prototype in Python (3.8).
The algorithm is implemented using the framework jMetalPy\cite{Benitez19} and TensorFlow
2.5.0. Transformations were generated using OpenCV 4.5.2. NSGA-II parameters  were as follows: population size of $50$ individuals, a mutation rate of  20 \%, a
crossover rate of  80 \%. $200$ iterations (resp. $100$) were used when calculating with NC (resp. DSA) in order to have similar execution time. $5$ steps were used in total (\ie $5$ reruns of the NSGA-II). \approach  has three mutation operators; we set the following mutation operator selection probabilities. Probability of changing value being $0.7$; the probability of nullifying the relation being $0.2$ and the probability of re-initializing the relation being $0.1$. Values were picked empirically to favour exploring neighbourhood rather than restarting the
exploration.
Implementations of the compared algorithm are the implementations provided by each respective papers with respective default parameters.

\begin{table}[!t] 
\caption{Transformations used in our experiments with parameters range. $[a, b]^2$ means transformation takes two parameters}
\renewcommand{\arraystretch}{1.2}
\centering
\begin{tabular}{c|c||c|l}
\hline
Rotation & $[-10, 10]$ & Shear & $[-0.1, 0.1]^2$\\
\hline
Translation & $[-2, 2]^2$ & Blur & $[0, 1.5]^2$ \\
\hline
Scale & $[0.9, 1.1]^2$ & Contrast & $[1, 2]$\\
\hline
\end{tabular}
\label{transf}
\end{table}


\section{Results}\label{results}
We now present the results obtained for each of our research questions. 
\subsection{RQ1: \rqone}
\textbf{RQ1}  deals with \approach performance versus a baseline of randomly generated HMR. More precisely, we examine the extent to which HMR built using HOMRS are better than randomly selected combinations of MR. 

\subsubsection{Criteria comparison : \approach vs random generation}

Comparison of criteria between random and best sets are presented in Table \ref{tab:res_first_exp_rand} for LeNet/MNIST and Table \ref{tab:res_first_exp_rand_vgg} for VGG/SVHN, with statistical test results presented in Table \ref{tab:res_first_exp_rand_stats} and Table \ref{tab:res_first_exp_rand_vgg_stats}. Note that for random sets on MNIST/LeNet, three sets had at least one HMR with uncertainty below threshold and thus were discarded.

\begin{table}
\caption{Comparison of the Effectiveness of \approach versus  Randomly Sampled Relations
  Set on LeNet/MNIST. Standard Deviation is Shown in Between parenthesis.}
\renewcommand{\arraystretch}{1.2}
\centering
\begin{tabular}{c|c|c|l}
\hline
 Calibration set & Cov  & Sim  & KR\\
\hline
\hline
 HOMRS (NC) & \textbf{82.87\%} & \textbf{86.00\%} & \textbf{76.35\%} \\
 & (0.0037) & (0.0085) & (0.0361) \\
\hline
 Random sets (NC) &  80.87\% & 94.08\% & 20.55\% \\
 & (0.0066) & (0.0147) & (0.1261) \\
\hline
\hline
HOMRS (DSA) & \textbf{67.79\%} & \textbf{81.01\%} & \textbf{65.46\%} \\
 & (0.0037) & (0.0085) & (0.0361) \\
\hline
 Random sets (DSA) &  66.63\% & 85.53\% & 20.55\% \\
 & (0.0088) & (0.0089) & (0.0683) \\
\hline
\end{tabular}
\label{tab:res_first_exp_rand}
\end{table}

\begin{table}
\caption{Mann-Whitney p-Values and Cliff's Delta for comparison random/\approach on MNIST/LeNet. $H_0$: \approach  Does not Provide Significant Better Relations Set Compared to Random Method.}
\renewcommand{\arraystretch}{1.2}
\centering
\begin{tabular}{c|c|c|l}
\hline
 & NCov  & NSim  & KR\\
\hline
 p-value & $1.82 \times 10^{-6}$ & $2.87 \times 10^{-9}$ & $2.87 \times 10^{-9}$\\
\hline
 Cliff's delta & 1 (large) & -1 (large) & 1 (large) \\
\hline
\hline
 & DSA  & NSim  & KR\\
\hline
 p-value & $1.29 \times 10^{-3}$ & $5.46 \times 10^{-6}$ & $2.87 \times 10^{-9}$\\
\hline
 Cliff's delta & 0.66 (large) & -0.86 (large) & 1 (large) \\
\hline
\end{tabular}
\label{tab:res_first_exp_rand_stats}
\end{table}

\begin{table}
\renewcommand{\arraystretch}{1.2}
\centering
\begin{tabular}{c|c|c|l}
\hline
 Calibration set & Cov  & Sim  & KR\\
\hline
\hline
 HOMRS (NC) & \textbf{76.42\%} & \textbf{93.84\%} & \textbf{43.46\%} \\
 & (0.0031) & (0.0033) & (0.0339) \\
\hline
 Random sets (NC) &  75.61\% & 96.70\% & 18.47\% \\
 & (0.0031) & (0.0042) & (0.040) \\
\hline
\hline
HOMRS (DSA) & \textbf{70.79\%} & \textbf{85.81\%} & \textbf{39.11\%} \\
 & (0.0031) & (0.0064) & (0.0559) \\
\hline
 Random sets (DSA) &  69.99\% & 90.70\% & 18.47\% \\
 & (0.0053) & (0.0118) & (0.0488) \\
\hline
\end{tabular}
\caption{Comparison of the Effectiveness of \approach versus  Randomly Sampled Relations
  Set on VGG/SVHN.. Standard Deviation is Shown in Between Parenthesis.}
\label{tab:res_first_exp_rand_vgg}
\end{table}

\begin{table}
\renewcommand{\arraystretch}{1.2}
\centering
\begin{tabular}{c|c|c|l}
\hline
 & NCov  & NSim  & KR\\
\hline
 p-value & $5.29 \times 10^{-6}$ & $1.18 \times 10^{-9}$ & $1.18 \times 10^{-9}$\\
\hline
 Cliff's delta & 0.94 (large) & -1 (large) & 1 (large) \\
\hline
\hline
 & DSA  & NSim  & KR\\
\hline
 p-value & $2.36 \times 10^{-5}$ & $1.18 \times 10^{-9}$ & $4.72 \times 10^{-9}$\\
\hline
 Cliff's delta & 0.87 (large) & -1 (large) & 0.99 (large) \\
\hline
\end{tabular}
\caption{Mann-Whitney p-Values and Cliff's Delta for comparison random/\approach on SVHN/VGG. $H_0$: \approach  Does not Provide Significant Better Relations Set Compared to Random Method.}
\label{tab:res_first_exp_rand_vgg_stats}
\end{table}

It is clear from reported data that \approach performs much better than random HMR
generation no matter the criteria. If the difference isn't visible much on coverage, as it seems coverage has a tendency to saturate when a lot of data are used, the difference is visible in term of similarity and particularly in term of kill ratio. Comparing Neuron Coverage and Distance Surprise Adequacy in term of coverage and similarity isn't meaningful as they don't exactly measure the same thing. However, we can see an advantage in term of Kill Ratio when using Neuron Coverage. One could argue that Neuron Coverage usually leads to less "natural" images like it was pointed out in \cite{Harel20}, as such DSA could be more representative. However, we argue that the uncertainty threshold we impose ensures a certain consistency of the images generated. Moreover, we reduce the number of iterations of the optimization algorithm when using DSA, in order not to inflate too much the time complexity. Increasing the number of iterations to match the number used with NC coverage would probably enhanced the HMR sets obtained in term of criteria (see RQ3). 

\textbf{Regarding statistical significance}, all comparisons using Mann-Whitney test showed statistical significance ($p-$value $< 0.05$) which further highlight the performance of \approach over random sets. Effect sizes are in all cases \enquote{large}, even though it's less prevalent in the case of DSA, which could be explained by a lower number of iterations as we explained previously.

\subsubsection{Comparison of obtain sets when applied on calibration dataset and test dataset}

\begin{table}
\caption{Comparison of the criteria in between Calibration dataset and Test dataset when using our HMR sets on LeNet/MNIST.}
\renewcommand{\arraystretch}{1.2}
\centering
\begin{tabular}{c|c|c|l}
\hline
 & Cov  & Sim  & KR\\
\hline
\hline
 HOMRS / NC (cal. dataset) & 82.87\% & 86.00\% & 76.35\% \\
 & (0.0037) & (0.0085) & (0.0361) \\
\hline
 HOMRS / NC (test dataset) & 83.04\% & 85.86\% & 78.25\% \\
 & (0.0026) & (0.0069) & (0.0503) \\
\hline
\hline
 HOMRS / DSA (cal. dataset) & 67.79\% & 81.01\% & 65.46\% \\
 & (0.0037) & (0.0085) & (0.0361) \\
\hline
 HOMRS / DSA (test dataset) & 68.37\% & 80.80\% & 67.45\% \\
 & (0.0026) & (0.0069) & (0.0503) \\
\hline
\end{tabular}
\label{res_first_exp_test}
\end{table}

\begin{table}
\caption{Comparison of the criteria in between Calibration dataset and Test dataset when using our HMR sets on VGG/SVHN.}
\renewcommand{\arraystretch}{1.2}
\centering
\begin{tabular}{c|c|c|l}
\hline
 & Cov  & Sim  & KR\\
\hline
\hline
 HOMRS / NC (cal. dataset) & 76.42\% & 93.84\% & 43.46\% \\
 & (0.0031) & (0.0033) & (0.0339) \\
\hline
 HOMRS / NC (test dataset) & 78.45\% & 94.04\% & 41.25\% \\
 & (0.0032) & (0.0030) & (0.0329) \\
\hline
\hline
 HOMRS / DSA (cal. dataset) & 70.79\% & 85.81\% & 39.11\% \\
 & (0.0037) & (0.0085) & (0.0361) \\
\hline
 HOMRS / DSA (test dataset) & 74.33\% & 84.78\% & 37.36\% \\
 & (0.0026) & (0.0069) & (0.0503) \\
\hline
\end{tabular}
\label{res_first_exp_test_VGG}
\end{table}

When using \approach, we make an assumption that any results obtained on the calibration dataset will work on the test dataset. This comes from the fact that both dataset comes from the same input distribution. To verify that it holds, we compute criteria value on the same HMR sets obtained previously. Results can be found in Table \ref{res_first_exp_test} and \ref{res_first_exp_test_VGG}.

The difference between the two datasets for both criteria is relatively small ($<5\%$), which seem to indicate that the transformation learned on our calibration dataset does extend to the input distribution. 

\subsubsection{Uncertainty profile comparison: \approach vs random generation}

\begin{figure}
    \centering
    \begin{minipage}{0.5\textwidth}
        \centering
        \includegraphics[width=\textwidth]{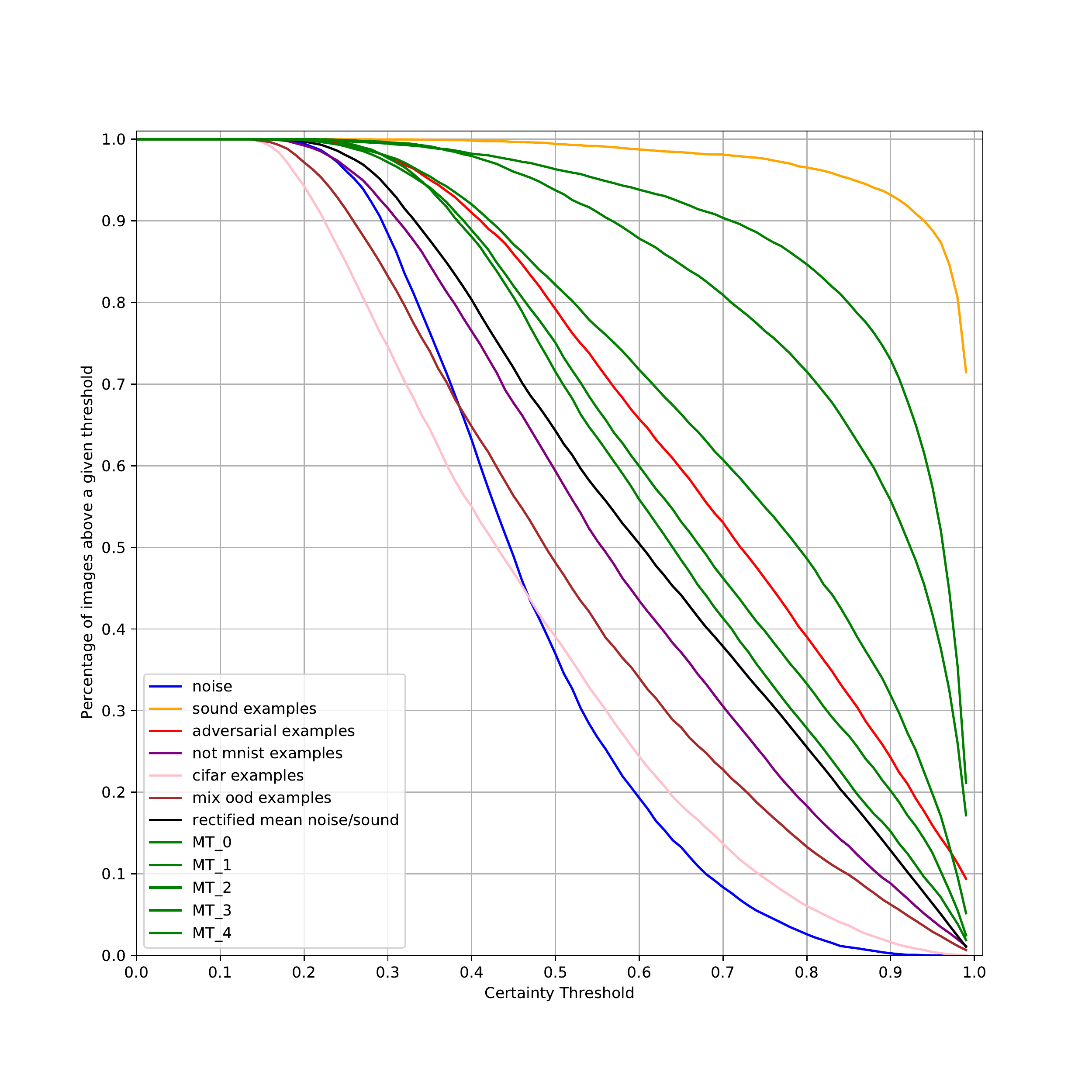} 
    \end{minipage}\hfill
    \begin{minipage}{0.5\textwidth}
        \centering
        \includegraphics[width=\textwidth]{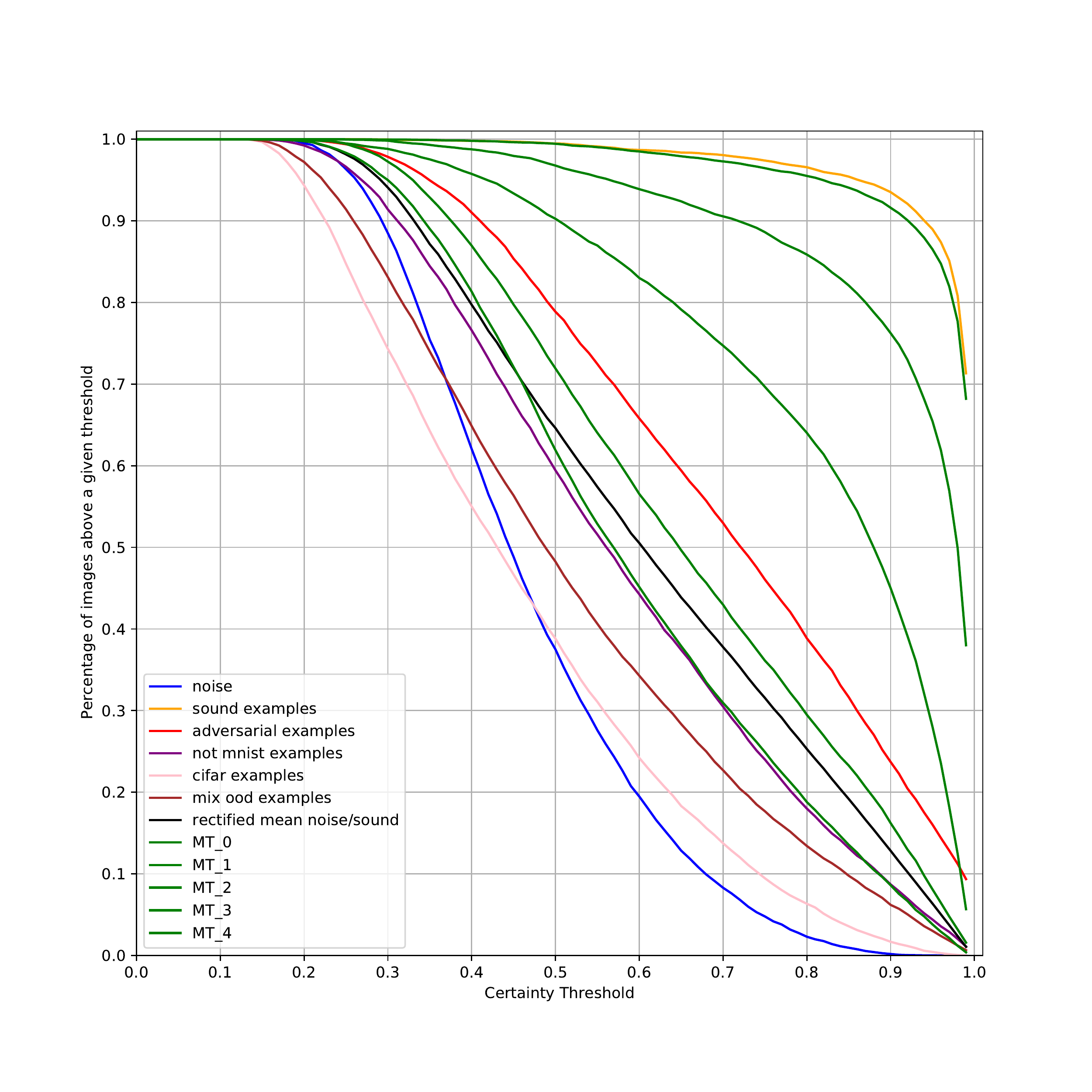} 
    \end{minipage}
    \caption{Uncertainty profile of HMR sets obtained by HOMRS (top) and randomly generated HMR (bottom) on LeNet/MNIST.}
    \label{fig_unc_profile_mnist}
\end{figure}

\begin{figure}
    \centering
    \begin{minipage}{0.5\textwidth}
        \centering
        \includegraphics[width=\textwidth]{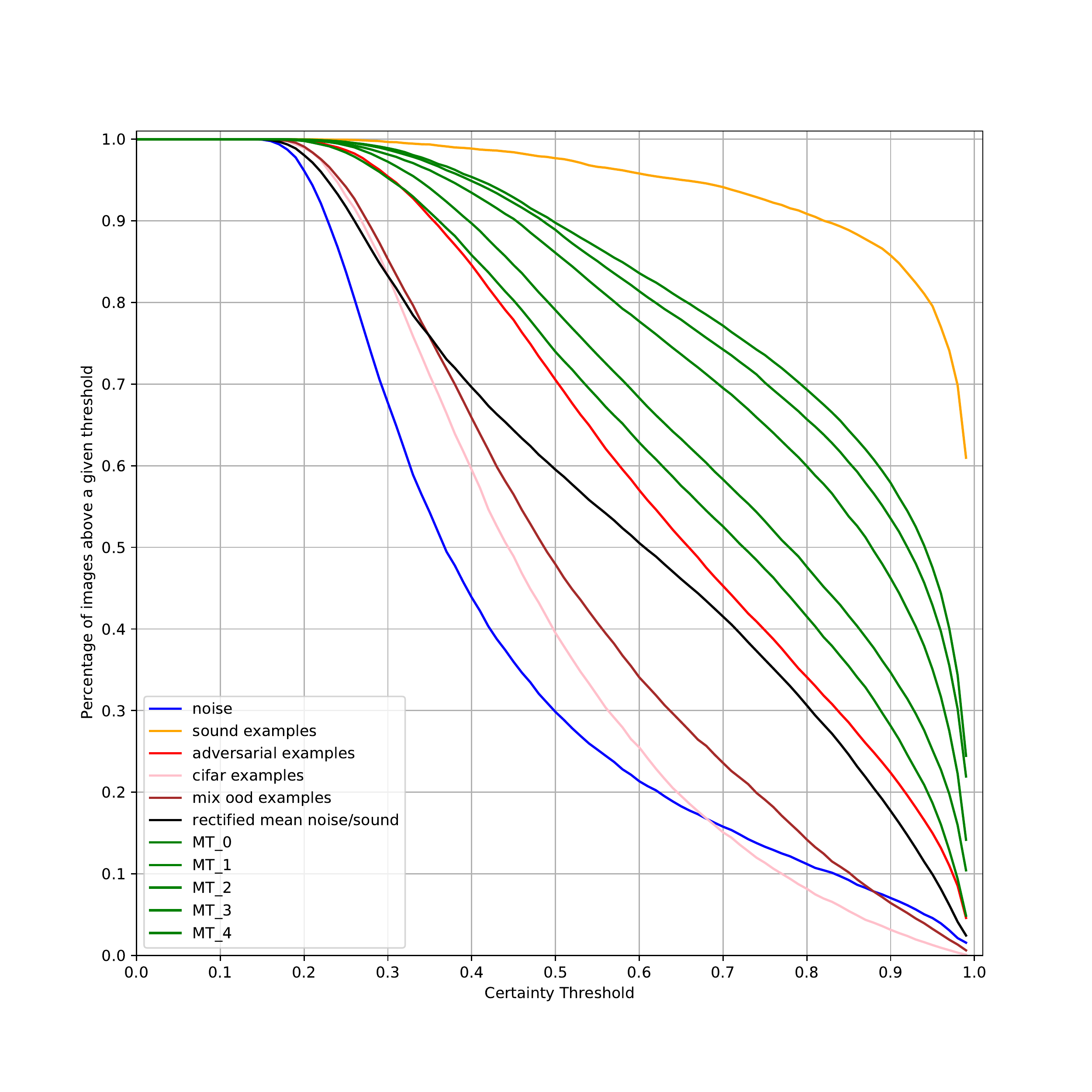} 
    \end{minipage}\hfill
    \begin{minipage}{0.5\textwidth}
        \centering
        \includegraphics[width=\textwidth]{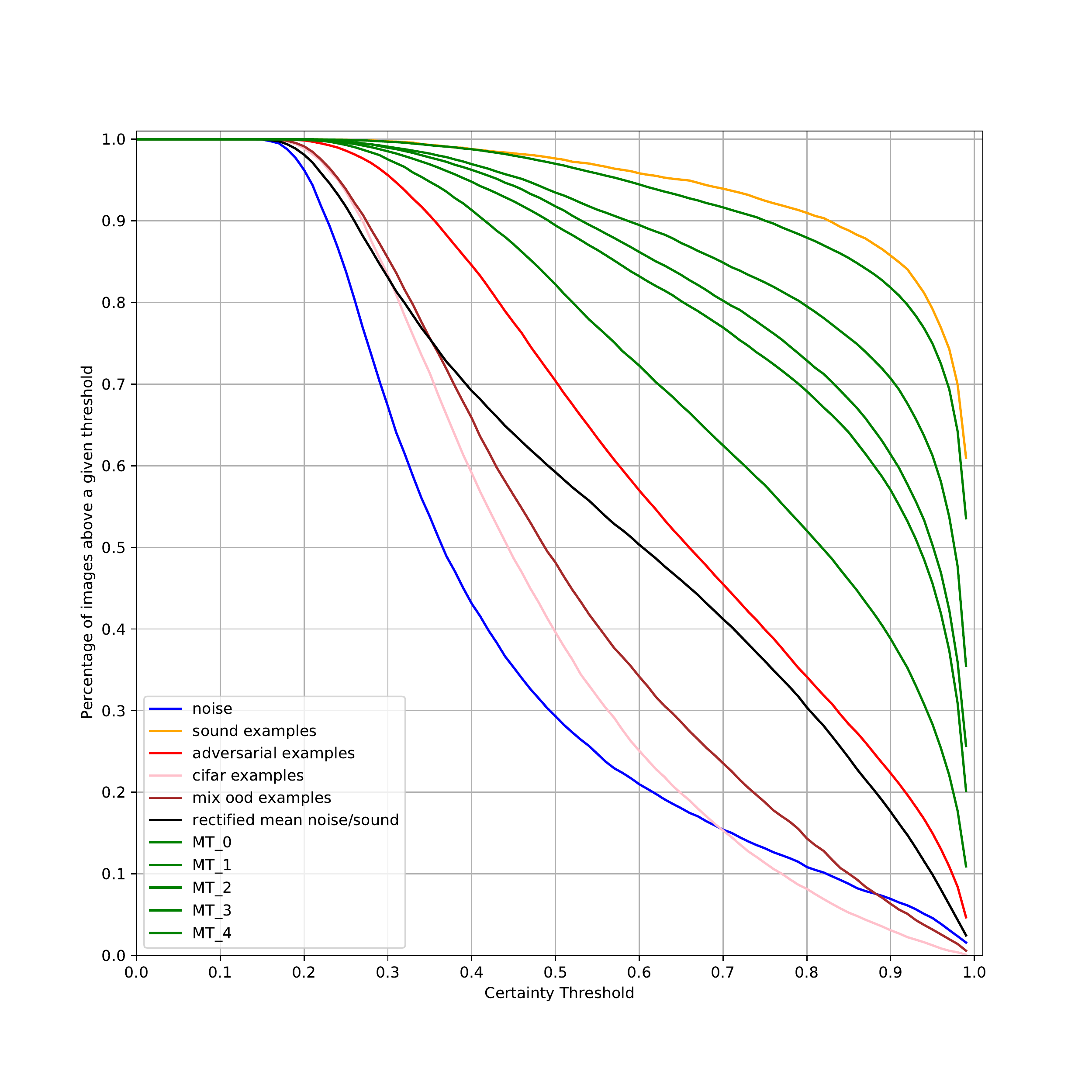} 
    \end{minipage}
    \caption{Uncertainty profile of HMR sets obtained by HOMRS (top) and randomly generated HMR (bottom) on VGG/SVHN.}
    \label{fig_unc_profile_svhn}
\end{figure}

We then investigated how the profile of an optimized HMR sets compare to a random sets. To do this, we provide uncertainty profile for two HMR sets (one optimized, one random). Similar behaviour can be observed on the other sets. We show the profiles on Figure \ref{fig_unc_profile_mnist} for MNIST/LeNet and on Figure \ref{fig_unc_profile_svhn} for VGG/SVNH.

In the case of MNIST/LeNet, we found some random sets containing one HMR which transformed distribution uncertainty profile is below our lower bound (and close to one OOD profile) which means images generated using this HMR lead to uncertainty distribution similar to one the model would expect from an OOD. This didn't happen in the case of our optimized sets. The random sets affected turned out to be the one yielding the highest Kill Ratio among random sets, which shows that high number of errors can be easily achieved if one doesn't control the validity of the transformation. As the transformation is not a valid transformation, the sets are discarded. As we are interested in transformation from the point of view of a distribution, we consider uncertainty as a distributional property, which means that a given transformation needs to behave properly for the given input distribution and model. 

For SVHN/VGG, no set (random or optimized) broke the uncertainty constraint, probably due to the more complex nature of the model/dataset (deeper network, more complex dataset, more training time...) and the basic MR parameters boundaries staying the same as with MNIST/LeNet. However one could see, as it is presented on Figure \ref{fig_unc_profile_svhn}, that random sets tend to have HMR with higher uncertainty profile, thus are less likely to probe the network for more erroneous behaviour, as we saw in Table \ref{tab:res_first_exp_rand_vgg}. Similar behaviour can be observed with the random sets presented for MNIST/LeNet, with notably one profile being very close to sound data.

\noindent\fbox{
\parbox{0.95\linewidth}{\textbf{RQ1 :} \textit{Relations obtained with HOMRS are more effective in term of metamorphic properties than randomly selected ones. Moreover, relations optimized on a calibration set does generalize on the input distribution. Images obtained with HOMRS are still relevant for the neural network as the distribution of the optimized transformations are above our empirical lower bound. This is not necessarily the case for randomly sampled sets, All of which highlight that careful selection of transformations is needed.}}} \\
\vspace{5pt}

\subsection{RQ2: \rqtwo}

\textbf{RQ2} aims at quantifying the impact of the hyper-parameters over \approach results. Three hyper-parameters were studied. Results when fixing $200$ evaluations can be found in Figure \ref{fig:hyp-eval} and \ref{fig:hyp-test}. Results when fixing $100$ and $400$ evaluations can be found in appendix.

\begin{figure}
    \centering
    \begin{minipage}{0.5\textwidth}
        \centering
        \includegraphics[width=\textwidth]{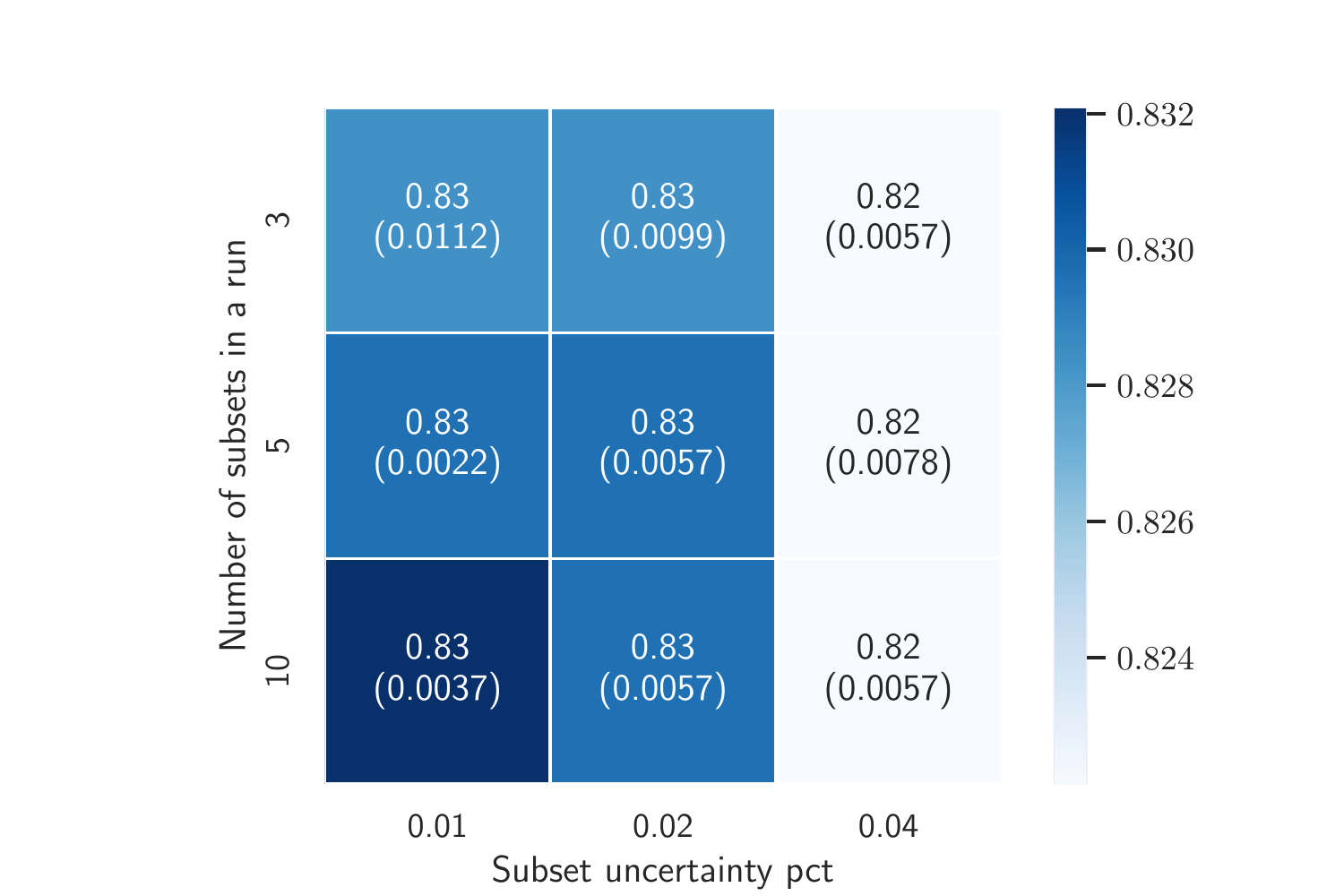} 
        \caption*{(a)}
    \end{minipage}\hfill
    \begin{minipage}{0.5\textwidth}
        \centering
        \includegraphics[width=\textwidth]{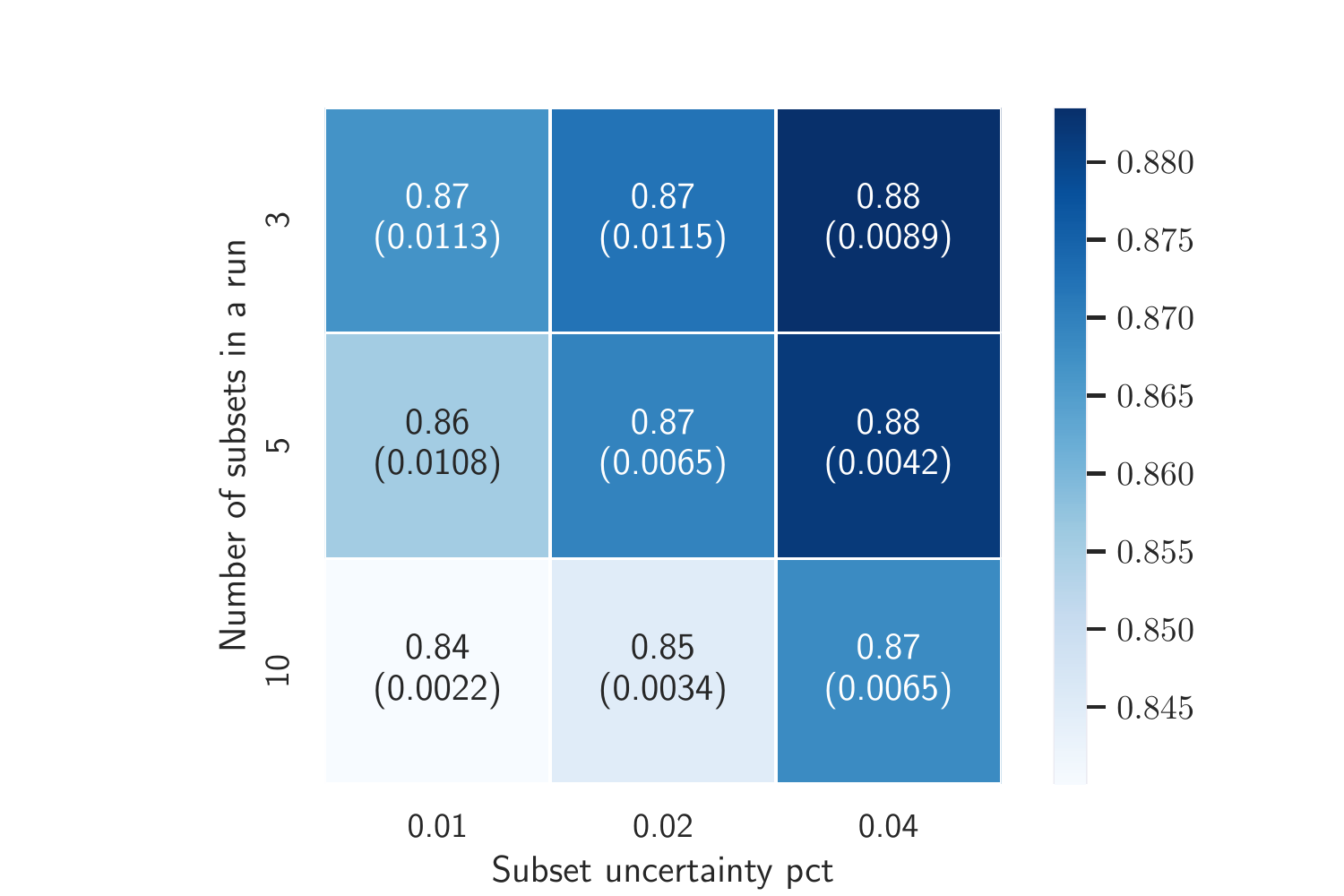} 
        \caption*{(b)}
    \end{minipage}
    \begin{minipage}{0.5\textwidth}
        \centering
        \includegraphics[width=\textwidth]{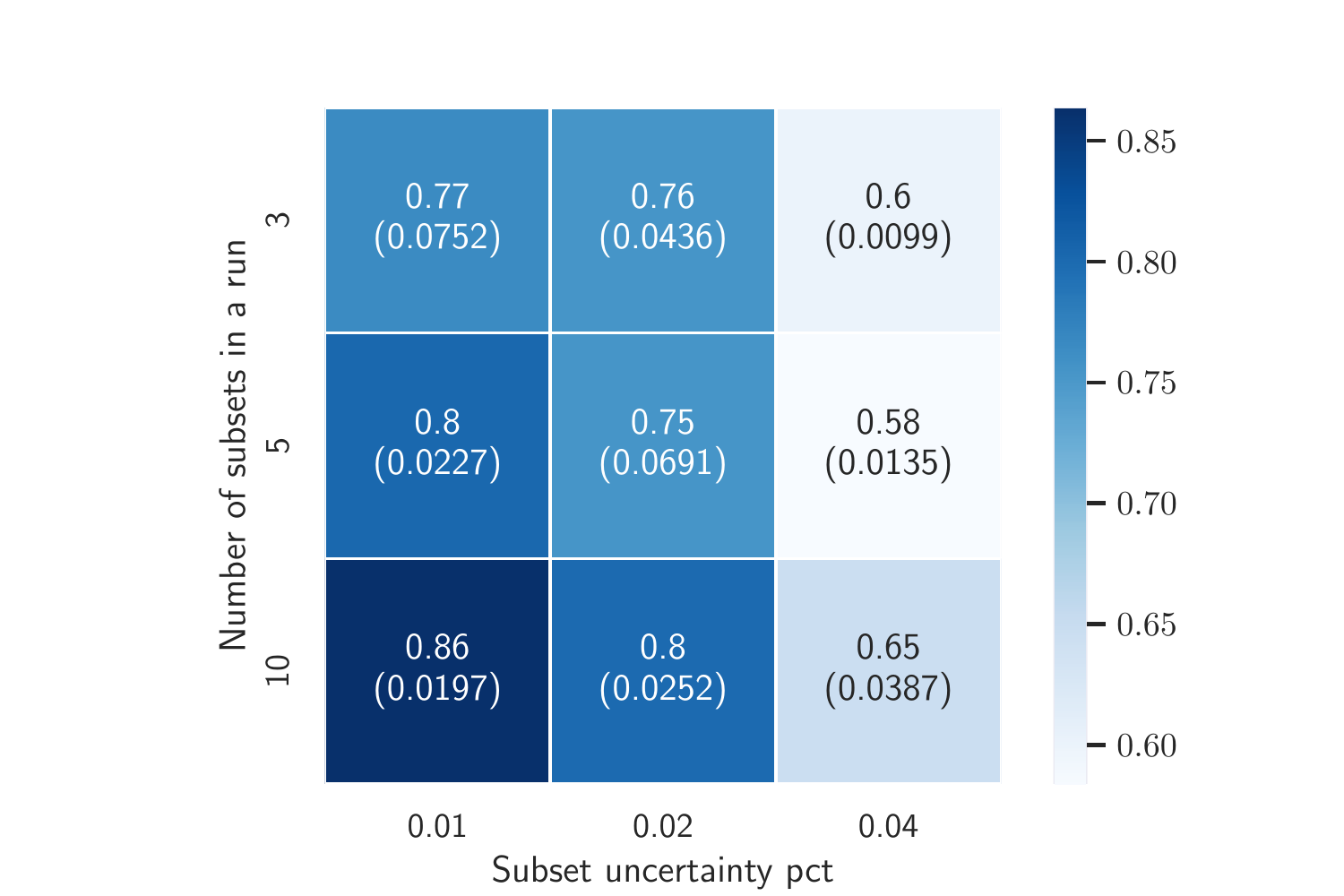} 
        \caption*{(c)}
    \end{minipage}
    \caption{Average over 3 independent runs for each of the criteria (Neuron Coverage (a), Similarity (b), Kill Ratio (c)) given $200$ evaluations when using the calibration set. Numbers in between parenthesis are the standard deviation.}
    \label{fig:hyp-eval}
\end{figure}

\begin{figure}
    \centering
    \begin{minipage}{0.5\textwidth}
        \centering
        \includegraphics[width=\textwidth]{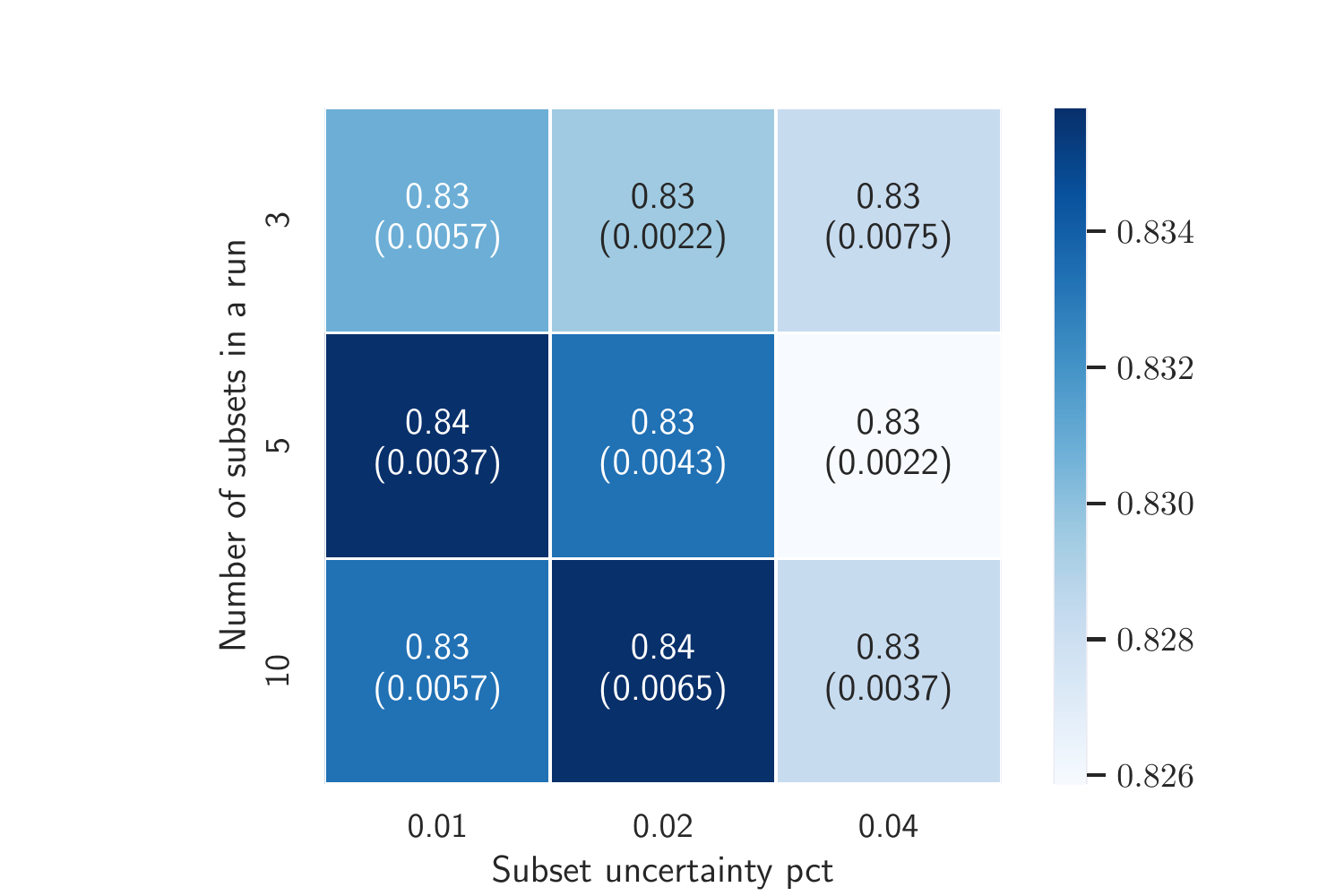} 
        \caption*{(a)}
    \end{minipage}\hfill
    \begin{minipage}{0.5\textwidth}
        \centering
        \includegraphics[width=\textwidth]{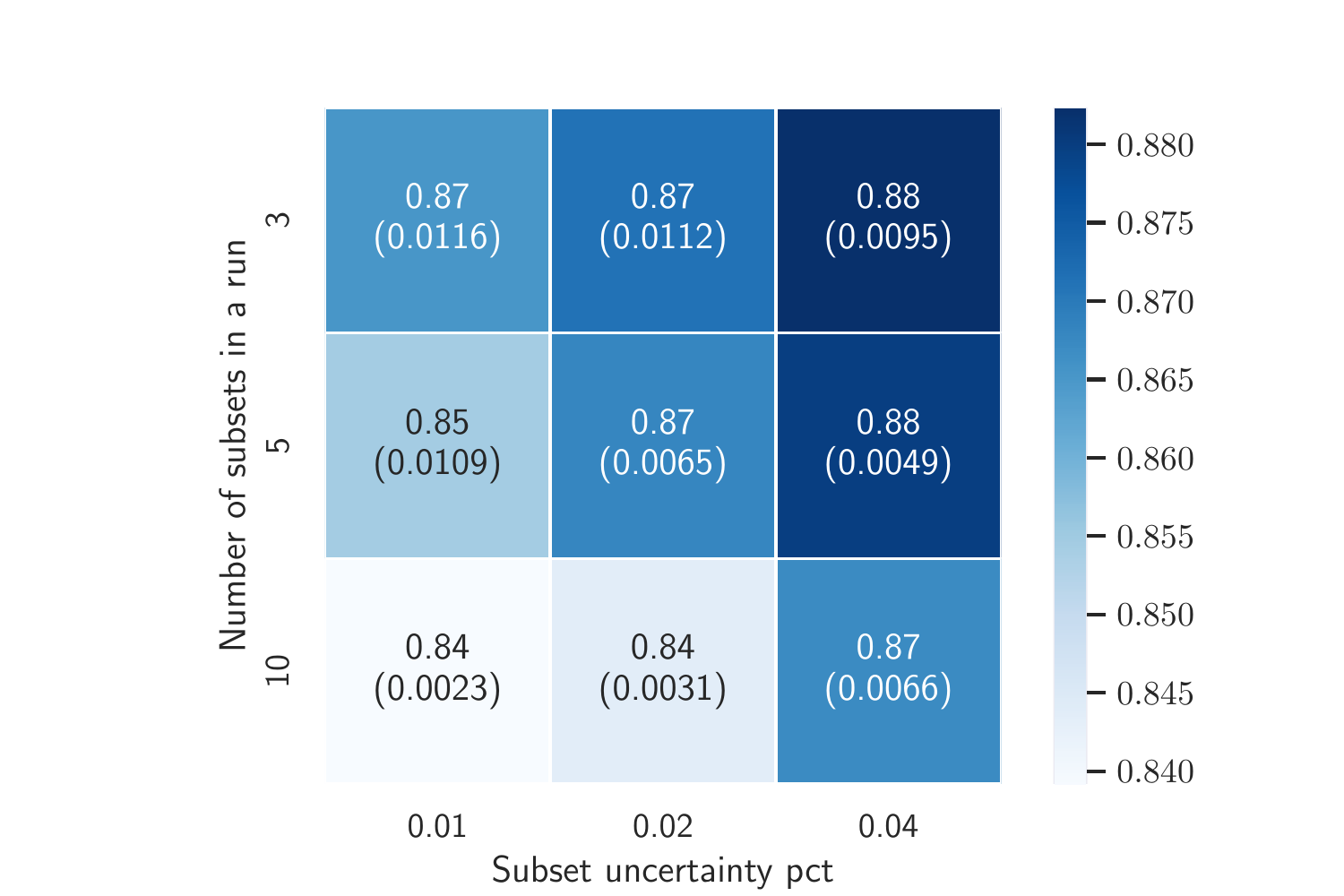} 
        \caption*{(b)}
    \end{minipage}
    \begin{minipage}{0.5\textwidth}
        \centering
        \includegraphics[width=\textwidth]{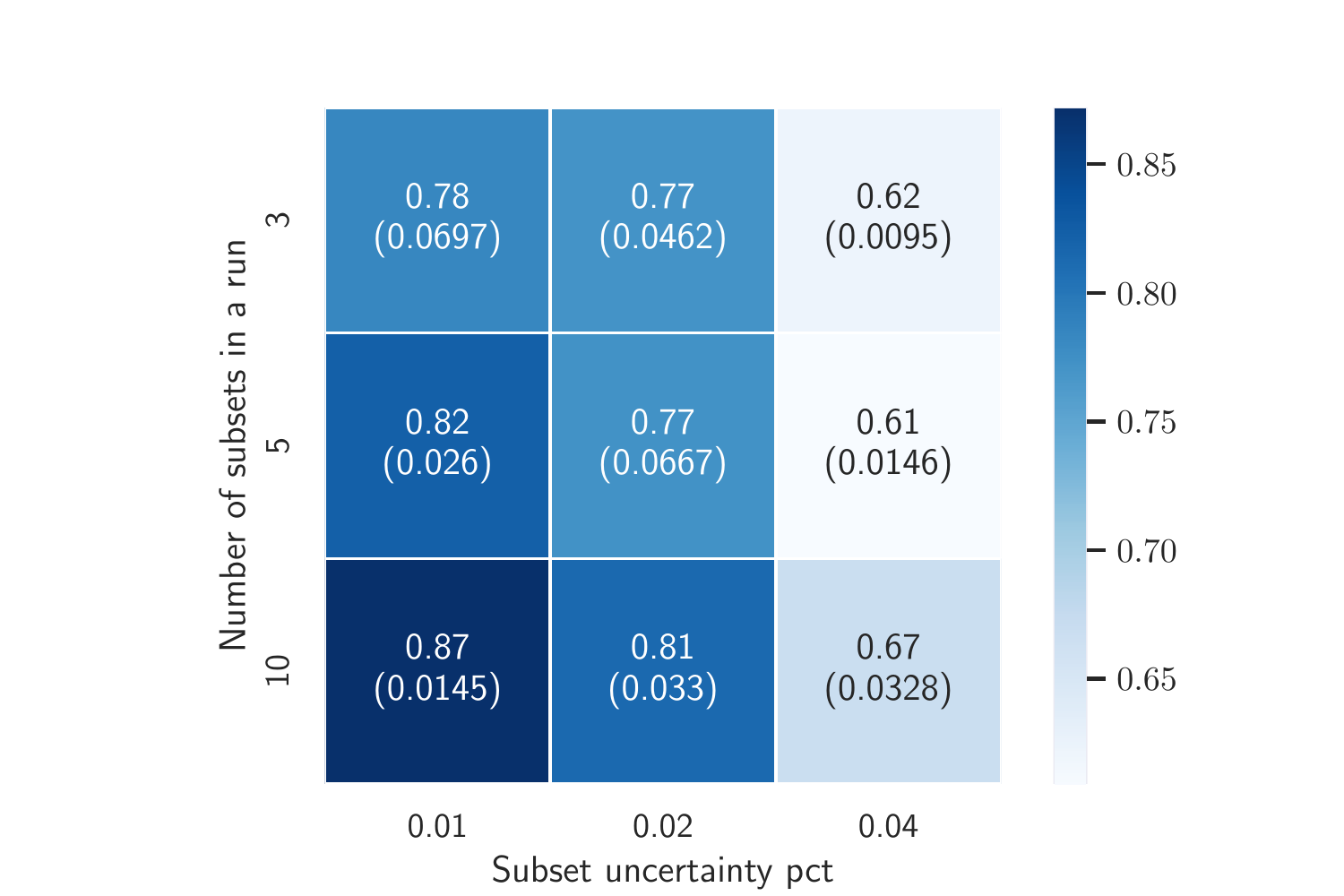} 
        \caption*{(c)}
    \end{minipage}
    \caption{Average over 3 independent runs for each of the criteria (Neuron Coverage (a), Similarity (b), Kill Ratio (c)) given $200$ evaluations when using the test set. Numbers in between parenthesis are the standard deviation.}
    \label{fig:hyp-test}
\end{figure}

As one can see on both figures, there is little change  between the calibration/test set for the same parameters, further emphasizing that the obtained HMR can generalize to any samples distribution extracted from the input distribution. Comparing with other number of evaluations (see appendix), it is clear that increasing the number of evaluations will have a tendency to generate HMR sets with increased Kill Ratio and decreased similarity, which seem logical as we leave the algorithm more time to explore the search space. However, there seems to be little effect on Neuron Coverage. For a fixed number of evaluations, it seems that decreasing the uncertain samples percentage injected in the subset used in the runs, as well as increasing the number of subsets used in a run, has a tendency to improve the quality of HMR sets w.r.t to criteria (increased Neuron Coverage/Kill Ratio and decreased Similarity). However, we observed that decreasing the uncertain samples percentage has a tendency to lower the obtained uncertainty profile obtained on the calibration and/or test set, meaning that we virtually obtain HMR sets that are not valid w.r.t to our definition of validity based on the empirical threshold we defined earlier. As such, to remain conservative on the uncertainty threshold, it might be better to increase the percentage of uncertain samples injected in the subset used in our runs. \\
\noindent\fbox{
\parbox{0.95\linewidth}{\textbf{RQ2 :} Increasing number of evaluations improve Kill Ratio and Similarity but have little effect on Neuron Coverage. The more subsets are used and the more uncertain samples are injected in subset for a run, the better HMR sets are in term of coverage, similarity and kill-ratio, yet increasing number of uncertain samples can results in lower then threshold uncertainty.}} \\

\subsection{RQ3: \rqthree}

\textbf{RQ3}  aims at quantifying the efficiency of \approach compared to other generation techniques both in term of coverage/adversarial examples generated as well as the validity of the examples generated.

\subsubsection{Comparison in term of criteria}

\begin{table}[H]
\caption{Comparison of the algorithm on LeNet/MNIST on the same subset. In the case of our algorithm, we average over the HMR sets found in previous experiment. We also provide execution of HOMRS and Random sets on the full test set.}
\renewcommand{\arraystretch}{1.2}
\centering
\begin{tabular}{c|c|c|l}
\hline
 & NC & \#Adv\\
\hline
\hline
 HOMRS (ours) & \textbf{79.85}\% & \textbf{399} (\textbf{825}) \\
\hline
 DistAware & 74.6\% & 94 \\
 \hline
 DeepXplore & 75.40\% & 129 \\
 \hline
 DLFuzz & 76.10\% & 134\\
 \hline
 \hline
 HOMRS (full set) & \textbf{83.04}\% & \textbf{7,825} (\textbf{15,926}) \\
 \hline
 Random sets (full set) & 81.71\% & 2,175 (3,021)\\
 \hline
\end{tabular}
\label{res_comp_crit}
\end{table}

\begin{table}[H]
\caption{Comparison of the algorithm on VGG/SVHN on the same subset. In the case of our algorithm, we average over the HMR sets found in previous experiment. We also provide execution of HOMRS and Random sets on the full test set.}
\renewcommand{\arraystretch}{1.2}
\centering
\begin{tabular}{c|c|c|l}
\hline
 & NC & \#Adv\\
\hline
\hline
 HOMRS (ours) & \textbf{68.36}\% & 210 (391) \\
\hline
 DistAware & 60.30\% & 169 \\
 \hline
 DeepXplore & 61.60\% & 291 \\
 \hline
 DLFuzz & 67.10\% & \textbf{1,272}\\
 \hline
 \hline
 HOMRS (full set) & \textbf{78.45}\% & \textbf{10,738} (\textbf{20,355}) \\
 \hline
 Random sets (full set) & 77.80\% & 4,665 (7,863)\\
 \hline
\end{tabular}
\label{res_comp_crit_vgg}
\end{table}

Note that even though DeepXplore offers three transformations, we only show the transformations with the highest coverage/number of adversarial examples (in our case, \emph{blackout} for both models). Similarly for DistAware (occlusion for MNIST, blackout for SVHN). All methods were used with the same random seeds of size $500$. We did the same with our method, even though our method is tailored to be effective on a whole distribution. We evaluated whether examples were truly adversarial by checking the difference of prediction between the original and generated images. Results for our method is the average neuron coverage/ number of adversarial examples of the selected HMR sets we choose in previous part. Note that we reported, for our method, \textit{unique} adversarial examples (\ie{} following our KR criteria where we count as only one an adversarial example even if multiple HMR generate multiple adversarial examples from the same input), as well as the total number of adversarial examples generated (if we count all examples that all our HMR generate) in between parenthesis.

Results are presented in Table \ref{res_comp_crit} for MNIST/LeNet and in Table \ref{res_comp_crit_vgg} for SVHN/VGG. For MNIST/LeNet. We see that, on the seed of siwe $500$, in terms of Neuron Coverage, \approach is better than the three other methods, improving on average by $3\%$ in comparison to DLFuzz. Moreover, our method manages to produce on average more than two times the number of adversarial examples of other methods, when considering only one HMR per input, but more than six times the number of adversarial examples of other methods, if we consider multiple HMR transformations per input. For SVHN/VGG, while \approach is still better, the improvement over the second best, DLFuzz, is on average only $1\%$. In terms of generated adversarial examples, DLFuzz generates roughly three times \approach total of generated examples, \approach generating itself more than two times DistAware and one and half times DeepXplore number of generated examples. The difference between MNIST/SVHN in our case may be explained by the limited parameter range of the transformation as well as the VGG model being more robust to transformations such as translation, rotation...etc. This, added to the limited seed size that is detrimental for our method (as it benefits from a higher number of inputs) could explain the difference obtained between the models. 

We also reported the numbers obtained on the full dataset as a comparison, for the random HMR sets as well as our optimized sets. This allows us to show the usefulness of selecting transformations that can be generalized. Once the HMR are optimized on the calibration sets, they can be applied on any new data, as long as the data come from the same input distribution. Hence, contrary to other methods for which time complexity depends on the size of the seeds and which need to be re-run to generate new examples, we can apply our method immediately on any number of given data once calibration is done. Moreover, this calibration is fast as it is done on small subsets based on the calibration set. Moreover, we can see that even using simple random sets in combination with a high number of data can already yield better results than the compared methods used on smaller seeds.

\subsubsection{Comparison in term of uncertainty}

\begin{figure}
    \centering
    \begin{center}
    \includegraphics[width=.45\textwidth]{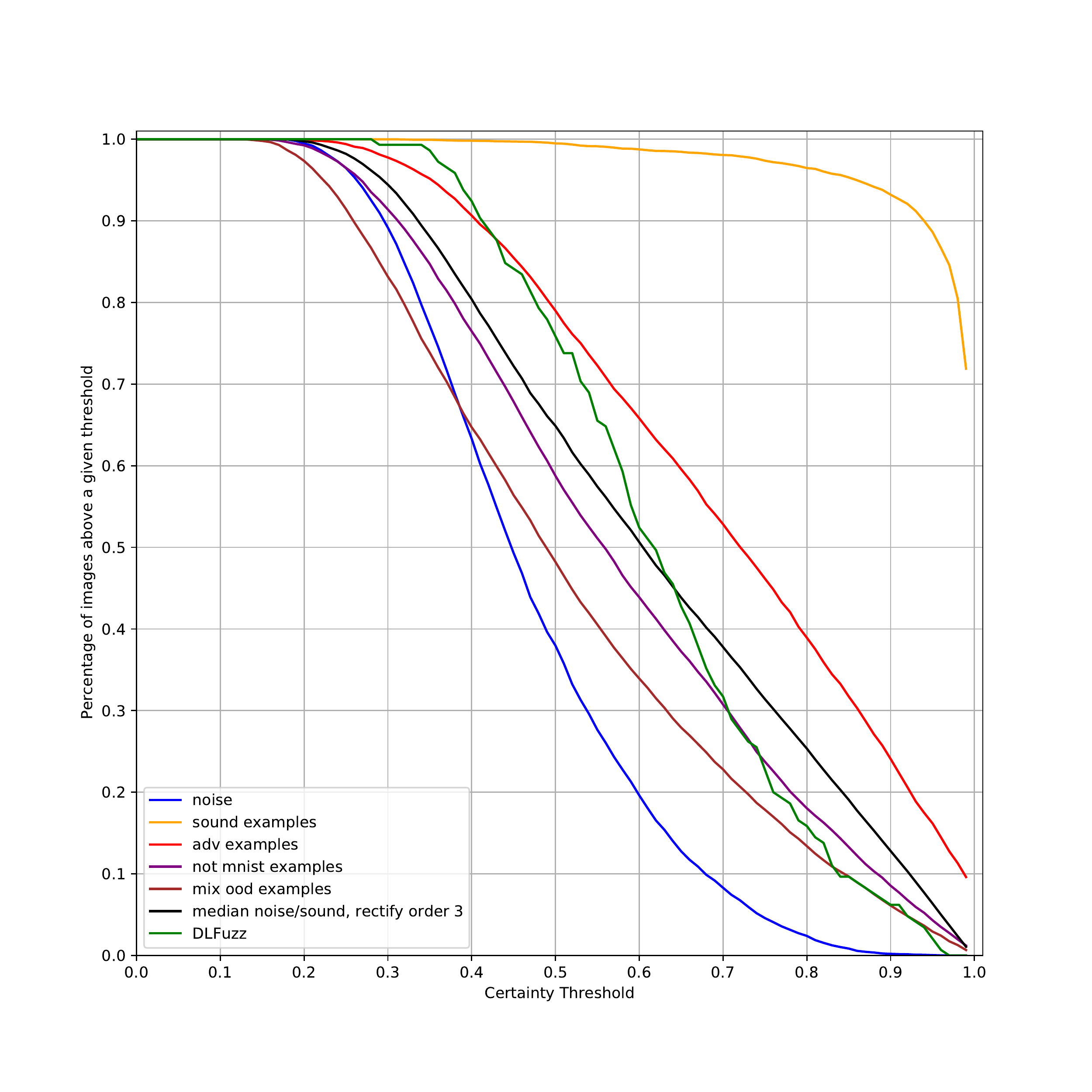}
    \includegraphics[width=.45\textwidth]{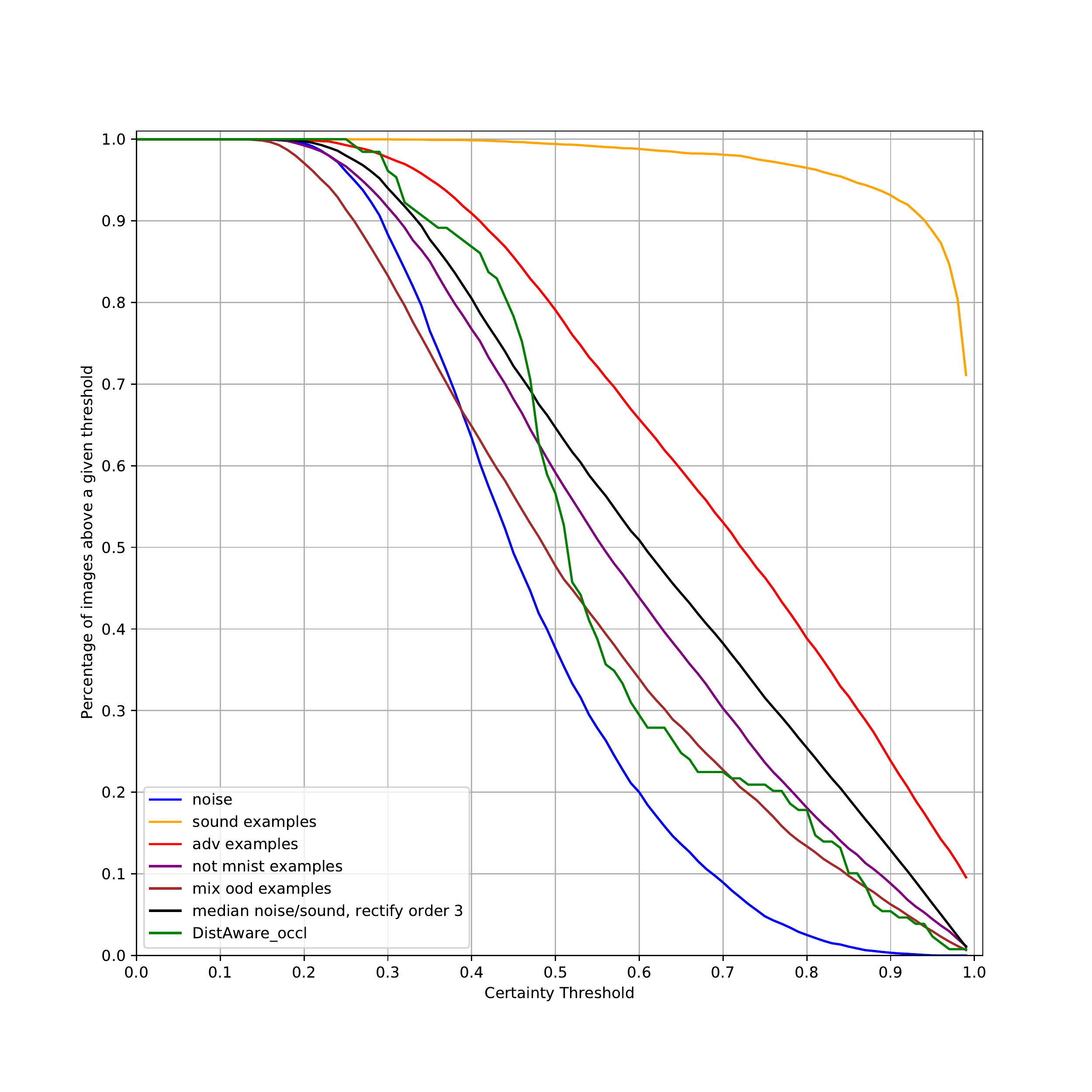}
    \includegraphics[width=.45\textwidth]{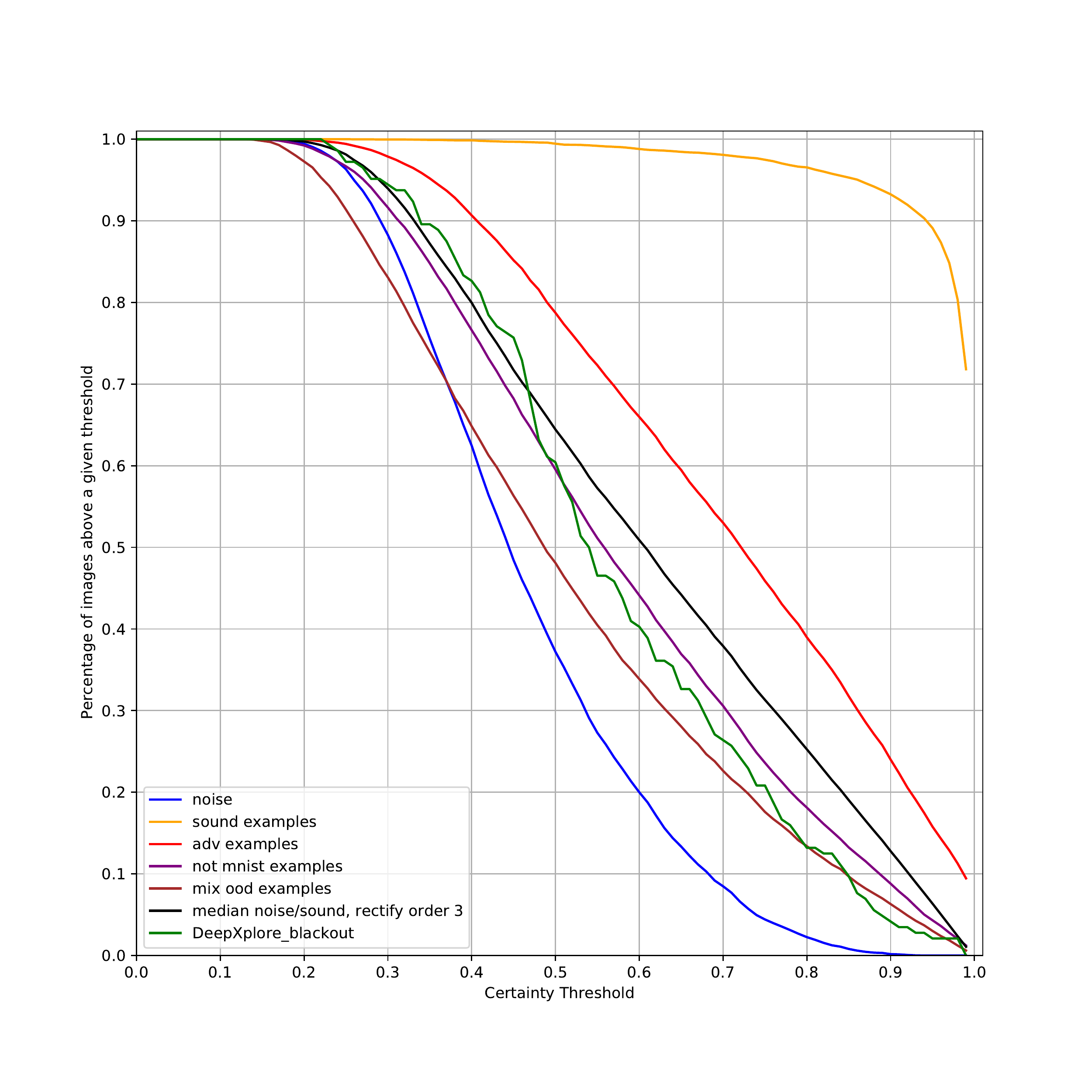}
    \includegraphics[width=.45\textwidth]{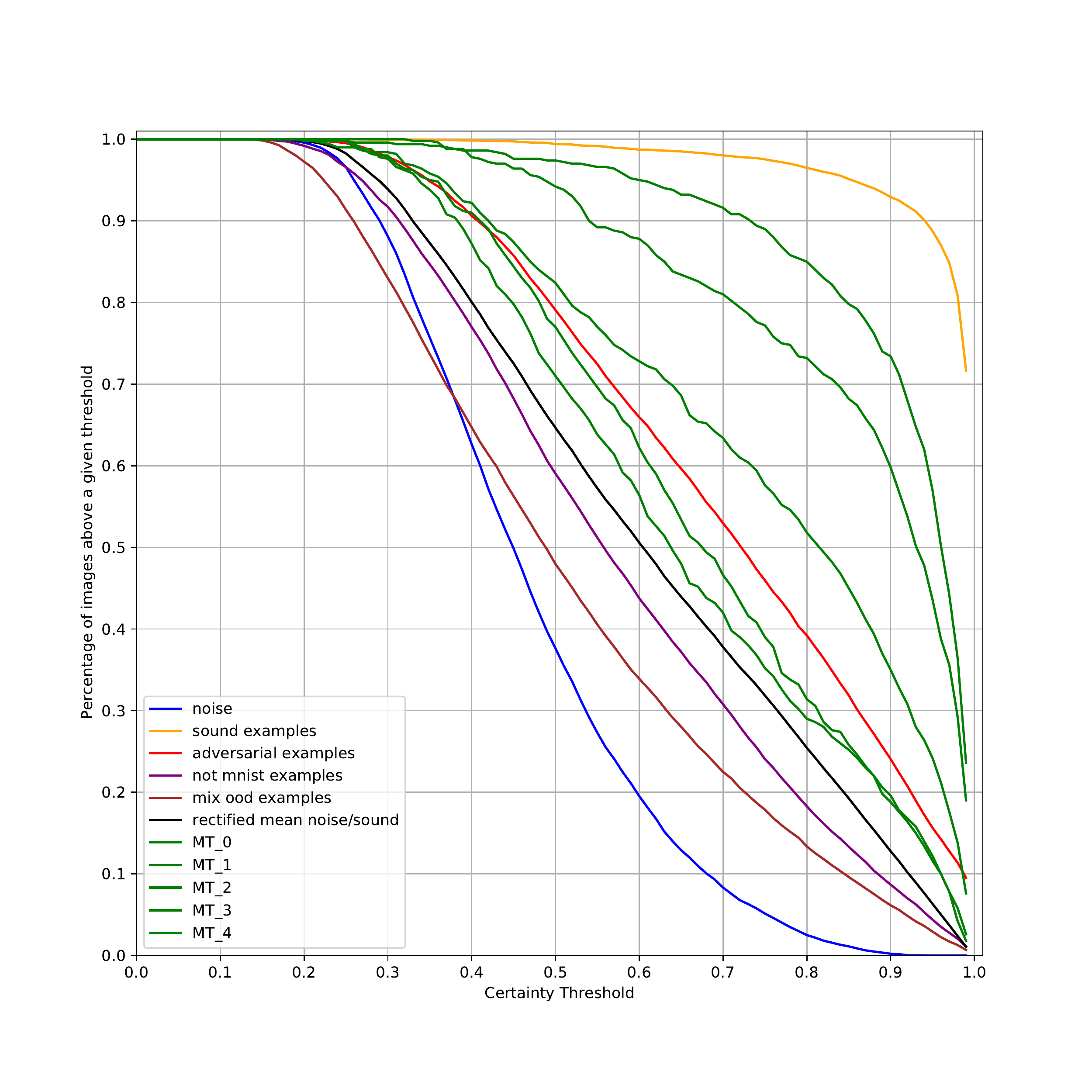}
    \end{center}
    \caption{Uncertainty Profile for MNIST/LeNet of (top to bottom and left to right) DLFuzz, DistAware, DeepXplore and HOMRS (ours). Uncertainty of the method is in green.}
    \label{fig:fig_comp_unc}
\end{figure}

In the previous parts, we evaluated the different methods in terms of Neuron Coverage and number of Adversarial examples. We will now evaluate them in terms of uncertainty profiles. In each case, we considered the generation method of each method as the transformation to be evaluated with an uncertainty profile. Similarly to our method, we gathered generated data and computed using MC Dropout, the uncertainty profile. For our method, we calculate the profile only using the same seed as other methods. Results for MNIST/LeNet are presented in Figure \ref{fig:fig_comp_unc} (in each case, the generated distribution is in green).

DLFuzz profile is partly under the curve, in the high certainty part, which also point out towards a seldom understanding of the generated distribution by the model, as such, the fuzzing process can be regarded as not entirely valid for the default parameters. Similarly, DeepXplore also is below the threshold curve. Regarding DistAware, one can see that the profile is largely under the threshold we computed. As such, it is interesting to see that, while all data generated through DistAware are supposed to be valid w.r.t the learned distribution of the VAE, the distribution of data generated isn't valid w.r.t to the uncertainty profile. Finally, \approach has uncertainty profile above the threshold even on the seed used for comparison, which further highlights that the method allows for a good generalization over the input distribution. Note that for our methods, we display separately all the different transformations found by our algorithm.

\begin{figure}
    \centering
    \begin{center}
    \includegraphics[width=.45\textwidth]{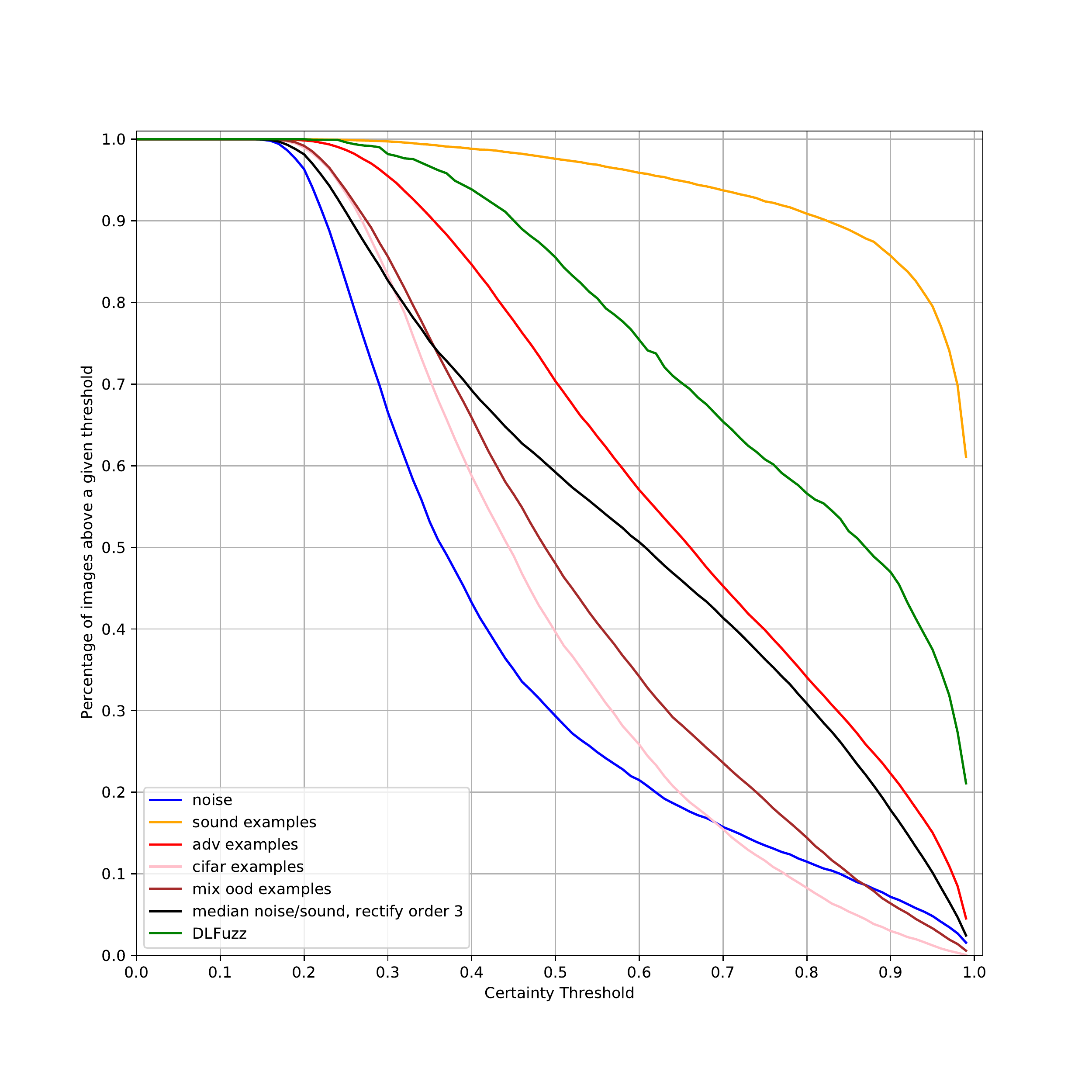}
    \includegraphics[width=.45\textwidth]{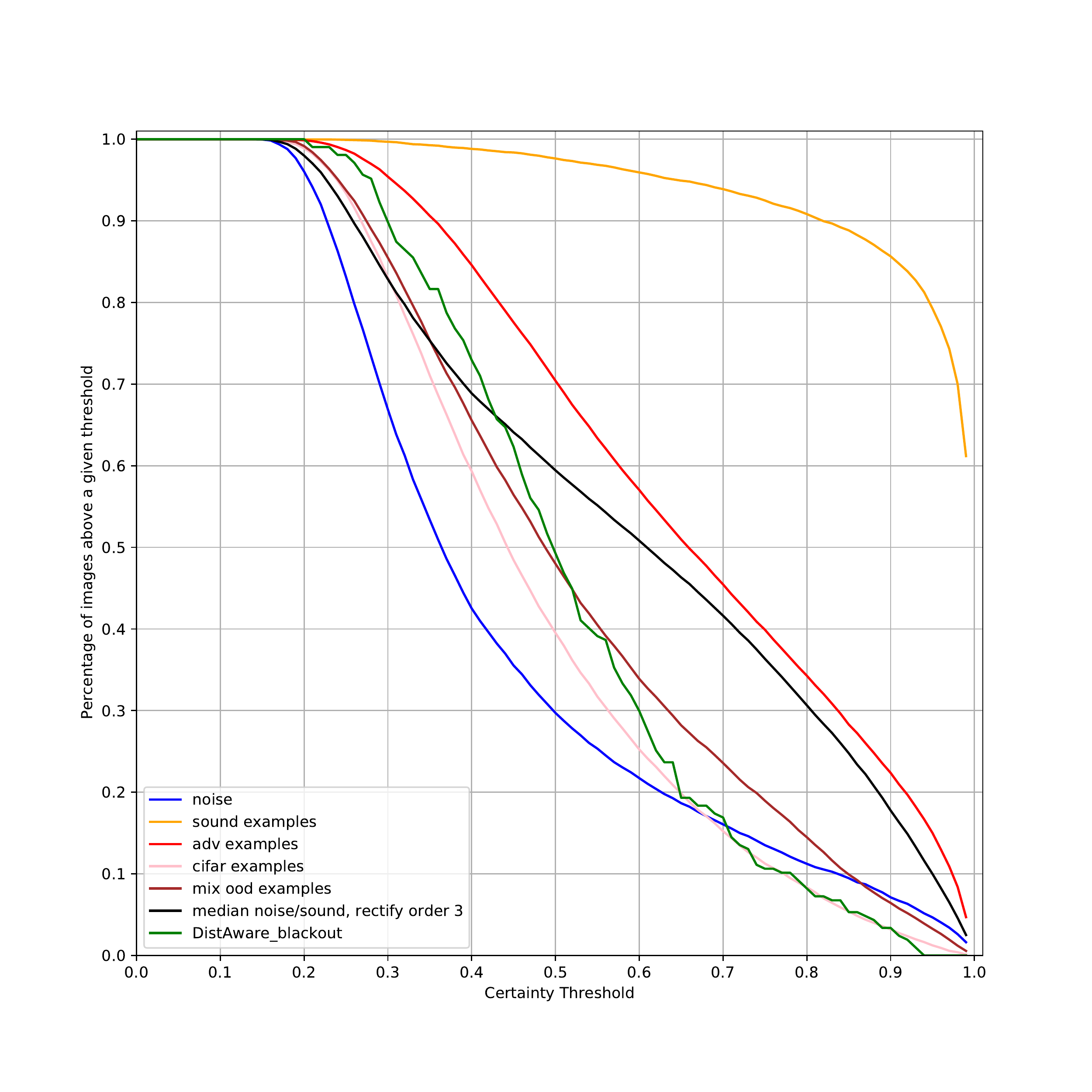}
    \includegraphics[width=.45\textwidth]{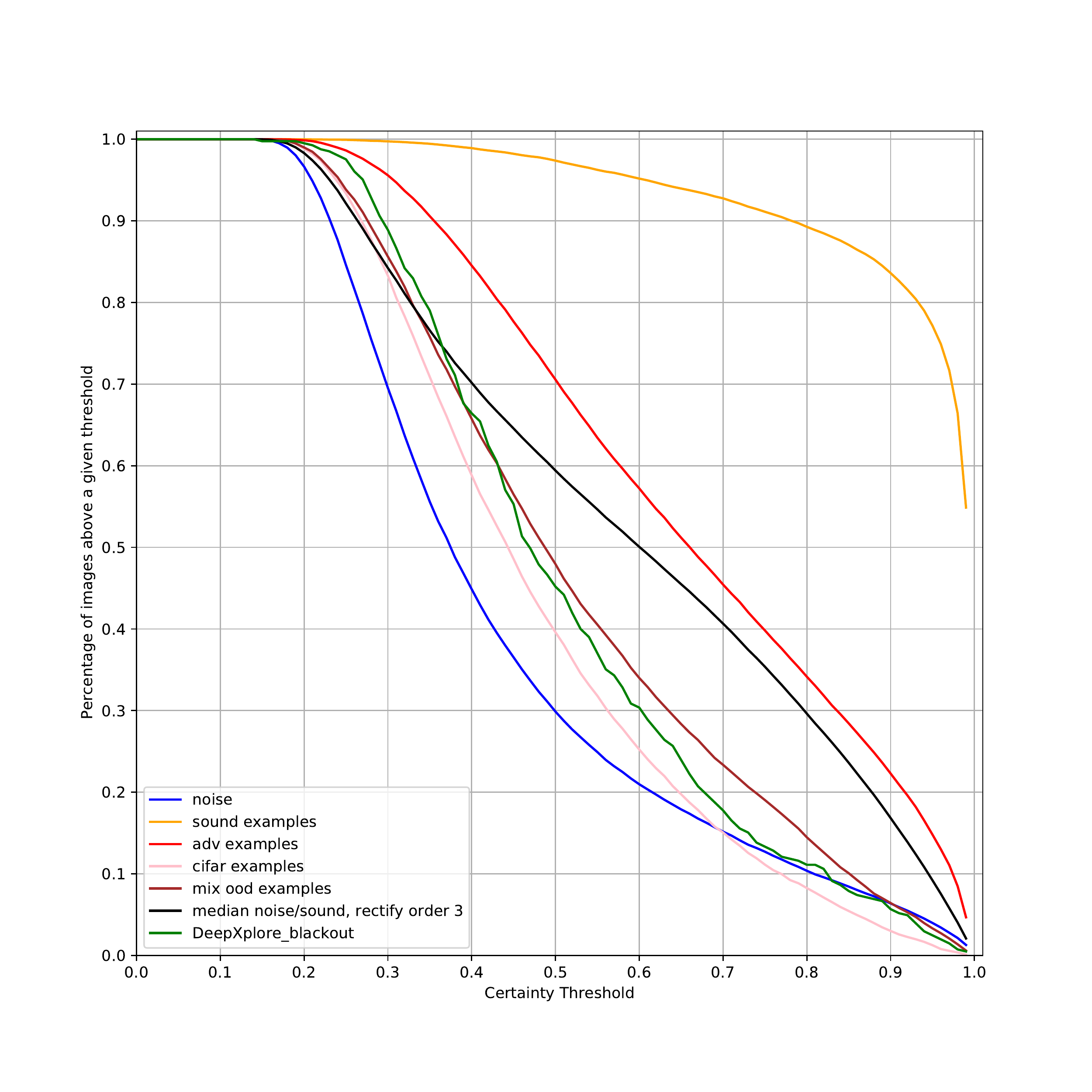}
    \includegraphics[width=.45\textwidth]{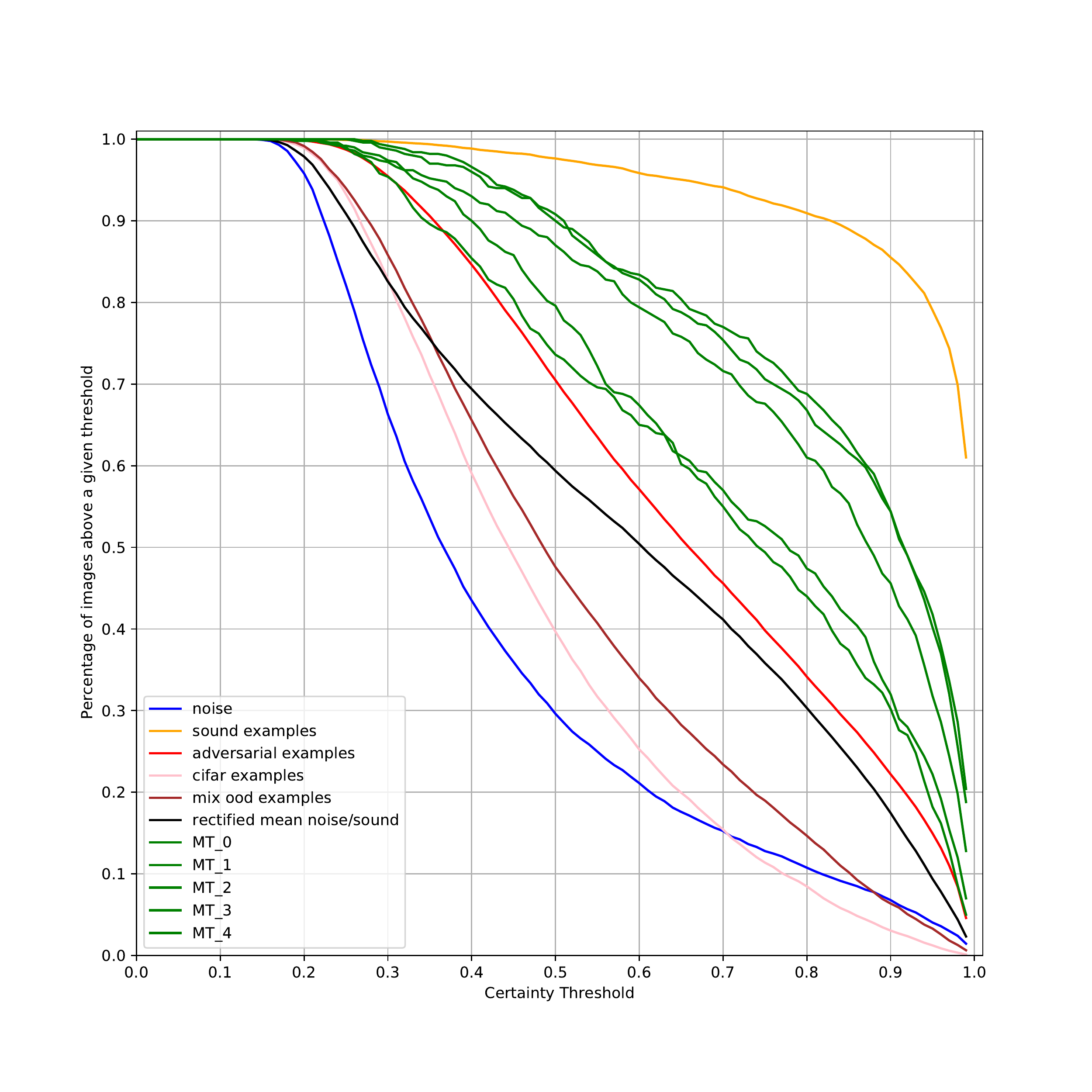}
    \end{center}
    \caption{Uncertainty Profile for SVHN/VGG of (top to bottom and left to right) DLFuzz, DistAware, DeepXplore and HOMRS (ours). Uncertainty of the method is in green.}
    \label{fig:fig_comp_unc_SVHN}
\end{figure}

Results for SVHN/VGG are presented in Figure \ref{fig:fig_comp_unc_SVHN} (in each case, the generated distribution is in green).
Here, DLFuzz is above the threshold (and above the curve of adversarial examples), similarly to our method. However, DeepXplore and DistAware both exhibit the same behavior as on MNIST/LeNet with their transformations being way below our empirical threshold.

\noindent\fbox{
\parbox{0.95\linewidth}{\textbf{RQ3 :} HOMRS generation methods result in a better adversarial examples generation and a good neuron coverage compared to similar methods. In particular, on similar seed size, HOMRS is the best in term of coverage on both model and the best in term of adversarial examples generated for MNIST/LeNet and second best on SVHN/VGG. Moreover, it's the only method among the compared one that have uncertainty threshold above the lower empirical bound on both MNIST/LeNet and SVHN/VGG. If we consider the possibility of applying on the whole test set, \approach is better than all other methods on all criteria in all cases at no extra cost.}} \\

\subsection{RQ4: \rqfour}

\textbf{RQ4}  deals with \approach time processing time versus DLFuzz on SVHN/VGG as it is the only method that beats our algorithm (i.e., HOMRS) on the sample seed comparison.

We first performed a sanity check to evaluate whether the changes we've made to DLFuzz by upgrading the code to meet the new version of the libraries affected the results. The average over the 4 runs gave us a coverage of around $66\%$ (compared to 67\% previously) and a number of generated images of $1,100$ (compared to 1,300 previously). The difference can be explained by potential non-determinism (GPU, libraries, algorithm, randomness...) as well as differences introduced in between the library versions. However, this difference remains sufficiently small. Hence, we consider that we didn't alter the overall working of the algorithm. We then averaged the time of computation for our method (10 runs) and DLFuzz (4 runs). Results can be found in Table \ref{tab:time}.

\begin{table}[H]
\caption{Computing time in seconds for both our method (average over the 10 runs we did previously) and DLFuzz used on the seed of size $500$ (averaged over 4 runs).}
\renewcommand{\arraystretch}{1.2}
\centering
\begin{tabular}{|c|c|}
\hline
 HOMRS (ours) & DLFuzz\\
\hline
\hline
 $\sim$ 18,000 &  $\sim$ 60,000\\
 \hline
\end{tabular}
\label{tab:time}
\end{table}

As one can see, our algorithm is approximately 3 times faster than DLFuzz. The fact that HMR can be applied by batch to the subset we are searching on, instead of multiple iterations on a single input decreases the time it takes. Moreover, as the calibration procedure is independent from the seed size used in the generation steps, the difference will only increase as seeds of increasing size would be used as input.

\noindent\fbox{
\parbox{0.95\linewidth}{\textbf{RQ4 :} HOMRS is approximately 3 times faster than DLFuzz, and the difference can only get bigger as the seed size increases. Indeed, while HOMRS procedure is independent from the seed size used in the generation, DLFuzz processing time depends on the size of the input seed.}} \\


\section{Discussion}\label{discussion}
\approach was designed to optimize a set of base MR with given parameters bounds over a given DNN/dataset, generating HMR designed to maximize coverage, non-similarity and kill ratio, over the DNN/dataset. We believe that \approach is general enough as it is based  on composition of already known (or discovered) MR. Moreover, while in our experiments, base MR chosen are classical image transformations,  \approach aims to be versatile as the relations in the pool can be anything the user deems appropriate to test the model. The metamorphic properties of the relations are selected and validated by the user. We provided empirical evidence, in the case of the images transformations we used, that the relations used with the defined parameters range didn't alter them to the point where it wouldn't be useful for the DNN, using the notion of uncertainty profile over the transformed distribution. While given evidence are empirical, we showed that it is related to the notion of Adversarial Example (\ie examples in-distribution) and OOD (\ie examples out-of-distributions) with the empirical threshold we designed being a cut between those two distributions, hence acting as a pseudo \enquote{lower-bound} for our HMR sets (in order to design adversarial transformations). In particular, we showed that previous methods from the literature (i.e., DeepXplore, DLFuzz, and DistAware) doesn't 
necessarily meet these requirements; with generated distributions resulting in uncertainty profile similar to what one would expect from an OOD (showing that transformations generated by these methods might not be completely valid). In particular, we saw in RQ3 that, if DeepXplore results are not that surprising (as the constraint mechanism used to validate generated images is over the obtained gradient not the image itself), it is less clear w.r.t to DistAware and DLFuzz, as they displayed lower values than threshold profiles (in both case for DistAware, only on MNIST/LeNet for DLFuzz). Regarding DLFuzz, the difference between MNIST/LeNet and SVHN/VGG can be explained by the robustness of the model against the perturbation strength: in the second case, VGG, the model is more complex and required extensive training compared to LeNet, while we kept the same default parameters (so perturbation strength). As such, it is clear that the perturbation induced on MNIST led the model to more uncertainty than when dealing with VGG. This can be verified on \approach, where the uncertainty threshold across our HMR set were also lower. Regarding DistAware, it uses a VAE to learn from the training distribution and then compute a threshold based on OOD set. However, this can pose multiple problems: first, the VAE has its own approximation and the reconstruction is solely based on the training distribution. Secondly, the threshold is computed based on the F1 score between the input data and one OOD dataset. Alternative dataset as an OOD, while being clearly invalid data w.r.t to the input distribution, are way too different, meaning that the gap between the two distributions is pretty wide with a lot of alternative in between. To show this, we used adversarial examples used in the MNIST/LeNet to calculate which percentage the VAE would reject as invalid. Result showed that the model considered that over 90\% of the adversarial examples were regarded as invalid, a score similar to what was found on OOD. As such, it seems that the uncertainty metric, which probes model's knowledge based on the training distribution of its task, can be more effective at judging the validity of inputs.

One noticeable key advantage of \approach' HMR is that they benefit massively from generalization, as optimized transformations aim to be generalizable to any data from the input distribution. As such, once calibration is done, the input seed can be of any size without any major effect on the computation time. However, the compared algorithms don't benefit from this fact, which means that they are restricted by the input seed size. That is, if one computes results on a $500$ size seed and wants to generate more results, the re-computation on a new seed is needed. Moreover, expanding the size of the seed (or, in the case of some algorithms, increasing the number of runs...etc) in order to have a larger pool of potential data to generate from, will necessarily drastically increase the computation time as each input is mutated multiple times. We pointed out this fact through RQ4.

One potential limitation of \approach is that it requires the model under test to have been trained with Dropout, as it is necessary for MCDropout, and thus uncertainty quantification. However, we argue that this is a very small limitation, as Dropout was shown to improve generalization \cite{Srivastava14} and adding it to the architecture wouldn't be a drastic modification such as what using a full Bayesian Neural Network would be. 


\section{Threats to Validity}\label{threats}
We now discuss the threats to validity of our study following common guidelines for empirical studies \cite{yin2002applications}.

\textit{Construct validity threats} concern the relation between theory and observation. In this paper we made a few assumptions about DNN testing when defining quality criteria for MR. However, each of these assumptions is grounded in a theory that has been proven to be valid in the context of traditional software engineering testing, or in previous studies on DNNs. For example, our assumption that higher neuron coverage contributes to higher fault detection rates is shared by previous DNN testing techniques such as DeepTest\cite{Tian18} and DeepEvolution\cite{BenBraiek19}. The rationale behind this assumption is that the more the decision logic of a DNN is exercised, the higher are the chances of uncovering corner cases leading to the detection of errors. Note that if neuron coverage was mainly used, any coverage metric such as distance surprise adequacy, can be used. Should coverage on its own might have some inherent limitation, we add similarity and kill ratio criteria, along with validity constraint in order to enhance the generation and relevance of corner-cases.

\textit{Threats to internal validity} concern our selection of datasets, models, frameworks, and  analysis method. 
We mitigated this threat by selecting widespread datasets and models for our study. 
We also used popular frameworks to reduce the risk of computational errors in our implementations. For optimization, we implemented our algorithm based on NSGA-II which is widely used. We experimented with transformations that are commonly used by DNN studies in the computer vision domain. However, HOMRS can be easily applied to any kind of MR. The uncertainty based mechanism was used as validity metric which, while being empirically based, was checked against valid/invalid distribution. Yet, further proofs are needed to validate the concept, particularly concerning how the empirical threshold is defined.

\textit{Conclusion validity threats} concern the relation between the treatment and the outcome. We paid attention not to violate the assumptions of our algorithms and models. 

\textit{Reliability validity threats} concern the possibility of replicating this study. We have provided all the necessary details required to replicate our study. The datasets and models used are publicly available. In addition, we make our artifacts available
\footnote{\url{https://github.com/FlowSs/RepPackageHOMRS}}


\textit{Threats to external validity} concern the possibility to generalize our results. We have evaluated HOMRS on LeNet5 and VGG16 models using MNIST and SVHN dataset with commonly available relations. Nevertheless, further validation on different types of relations, and different model architectures are desirable, especially to study how the uncertainty empirical threshold behaves.



\section{Related Works}\label{related}
MR have been used in traditional software programming to test scientific software \cite{Kanewala19}, web services \cite{Sun11} and image software \cite{Just09} among others\cite{Segura16}. Selection or generation of MR applied to traditional software was also researched on such as using machine learning and graph-based representation relations to generate new ones \cite{Kanewala16} which restricts the application to programs where such a graph can be used. A concrete form to search for can be used with applications such as relations finding for trigonometry formulas \cite{Jie14}, which require to know what to look for. Category-choice frameworks can also be used \cite{Chen16}, but it requires to be able to make category-choice specification for a given problem.

It was then extended to application in Machine Learning and DNN. Basic relations for Machine Learning algorithms were first proposed in \cite{Xie11}. It was then applied to validate both machine learning and DNN such as CNN \cite{Ding17} or LSTM \cite{dwarakanath19} and in all the domains using DNN such as Natural Language Processing \cite{Pinjia20} or medical imagery \cite{Ding19}. However, not much research has been done on selection and/or generation of MR targeted to ML algorithm; indeed, methods for traditional software use Control-Flow Graph to derive relations which is not relevant to the case of DNN. We are only aware of a method \cite{spieker20} which uses a reinforcement approach through multi-arms bandit to select MR for a given problem based on the result of the prediction of the algorithm. Our work differs from their, both on the approach as we are using multi-objective genetic algorithm instead of a learned policy, as well as the selection process as we probe the DNN to get relevant neurons information instead of just exploiting the context information. Moreover, we deal with higher order relations through composition where they only work on elementary ones, while also checking the validity of obtained data.

MR were used as a tool for test generation through search-based methods from a single input; DeepTest\cite{Tian18} uses greedy search based on neurons coverage with driving scene images and DeepXplore\cite{Pei19} similarly uses neurons coverage but through a multi-models joint optimization. DeepEvolution\cite{BenBraiek19} also uses neuron coverage but through evolutionary algorithms. However, where they use MR for \textit{test cases generation}, we rather focus on MR \textit{selection} as we try to obtain the best set of relations for a given DNN. Even though the goal is not strictly to generate test cases, we do end up generating some through the application of the relations from our best set over the whole test set. The difference in approach lies in the consideration of application of the relations; where they rather apply multiple MR which increase coverage on a limited subset of samples, we try to find HMR that perform equally well on the whole dataset through optimization on a limited amount of subsets, in terms of coverage, similarity and errors finding. Moreover, those methods didn't, as we pointed out in the introduction, consider the validity of images. If they did, it was not transformation wise, which we did in our case as we aimed to provide transformation that can generalize.

\section{Conclusion}\label{conclusion}
In this paper, we presented \approach an approach to select a small set of  HMR in the context of DNN testing. The approach is based on multi-objective optimization with three distinct objectives inspired by  traditional software engineering and MT.
\approach aims at building a small set of HMR maximizing the neuron coverage and error detection while decreasing the similarity of HRMs generated, while making sure that generated HMR transformations are valid.

We report evidences showing the effectiveness of the  idea through experiments using MNIST/SVHN dataset. First, \approach was compared to a set of randomly generated HMR, outperforming it. We made sure HMR obtained generalized on the input distribution by testing them on both the calibration (validation) set and test set which yielded similar result. Similarly, we checked that uncertainty profiles remained above our threshold in all case. Secondly, we studied the set of hyper-parameters that are specific to \approach to analyze their impact on the HMR obtained. We showed that increasing the number of evaluation increases the quality of HMR set as the algorithm can explore the search space more thoroughly. We also showed that increasing the number of subsets improves the Kill Ratio and Similarity, yet at the cost of increased computation time. Analog observations can be made when decreasing the number of uncertain samples injected in the subset used by the NSGA-II. However, the later will have a tendency of potentially leading to HMR sets with uncertainty profile lower than the empirical threshold in some case, which is why we chose a higher value in our experiment in order to avoid this problem. Thirdly, we contrasted \approach with three coverage based generation methods: DeepXplore, DLFuzz, DistAware. When using the same seed of siwe $500$, \approach outperforms, in term of coverage and adversarial examples, all methods on MNIST/LeNet and is only bested by DLFuzz on SVHN/VGG in term of adversarial examples generated. However, the generalization property of \approach allows to apply, at not extra cost, obtained optimized HMR set to seed of any size. Hence, if we use as seed the whole test set, \approach outperforms all other methods on all accounts. Moreover, We also highlighted that not all of these methods necessarily respects the validity criteria that we established. Finally, we compared the computation time of DLFuzz and \approach on the same configuration. Our results showed that \approach is approximately three times faster when using the seed of size $500$. Moreover, as our approach can generalize easily once calibrated, this difference will only increase with an increasing seed size. 

MT is a powerful tool that can help test DNN. We believe that the \approach philosophy can stir the research in a new direction to tackle the MR selection problem. For future works, we aim to investigate the effectiveness of the method on more complex models and datasets, and improve on our approach, especially the uncertainty process in order to make it more grounded in theory and more robust.



\begin{acks}
This work was partially funded by NSERC through the DEEL project
\end{acks}

\bibliographystyle{ACM-Reference-Format}
\bibliography{bibfile}

\clearpage

\appendix
\section*{APPENDIX}

\section{Hyper-parameter tuning}

\begin{figure}[h]
    \centering
    \begin{minipage}{0.5\textwidth}
        \centering
        \includegraphics[width=\textwidth]{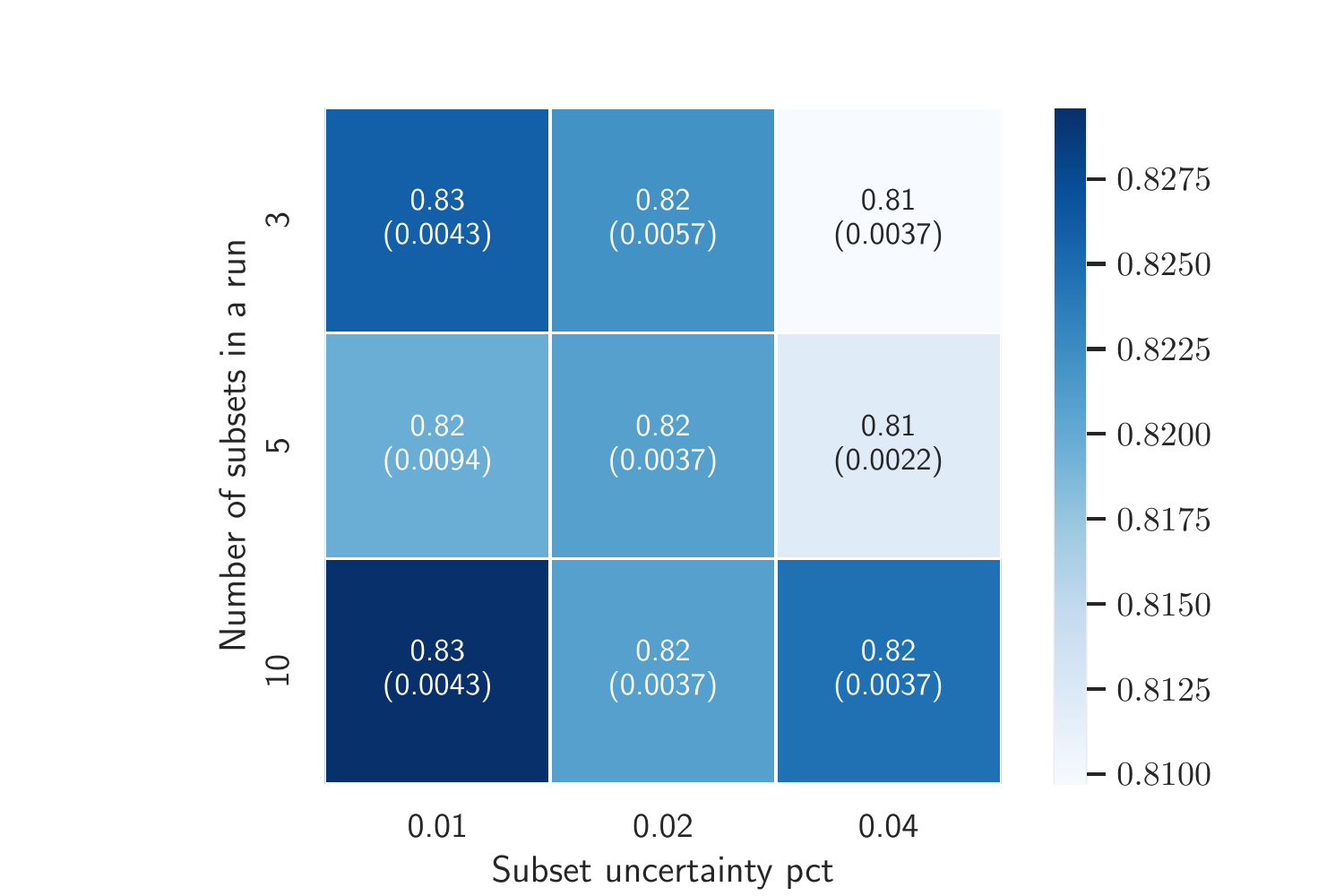} 
        \caption*{(a)}
    \end{minipage}\hfill
    \begin{minipage}{0.5\textwidth}
        \centering
        \includegraphics[width=\textwidth]{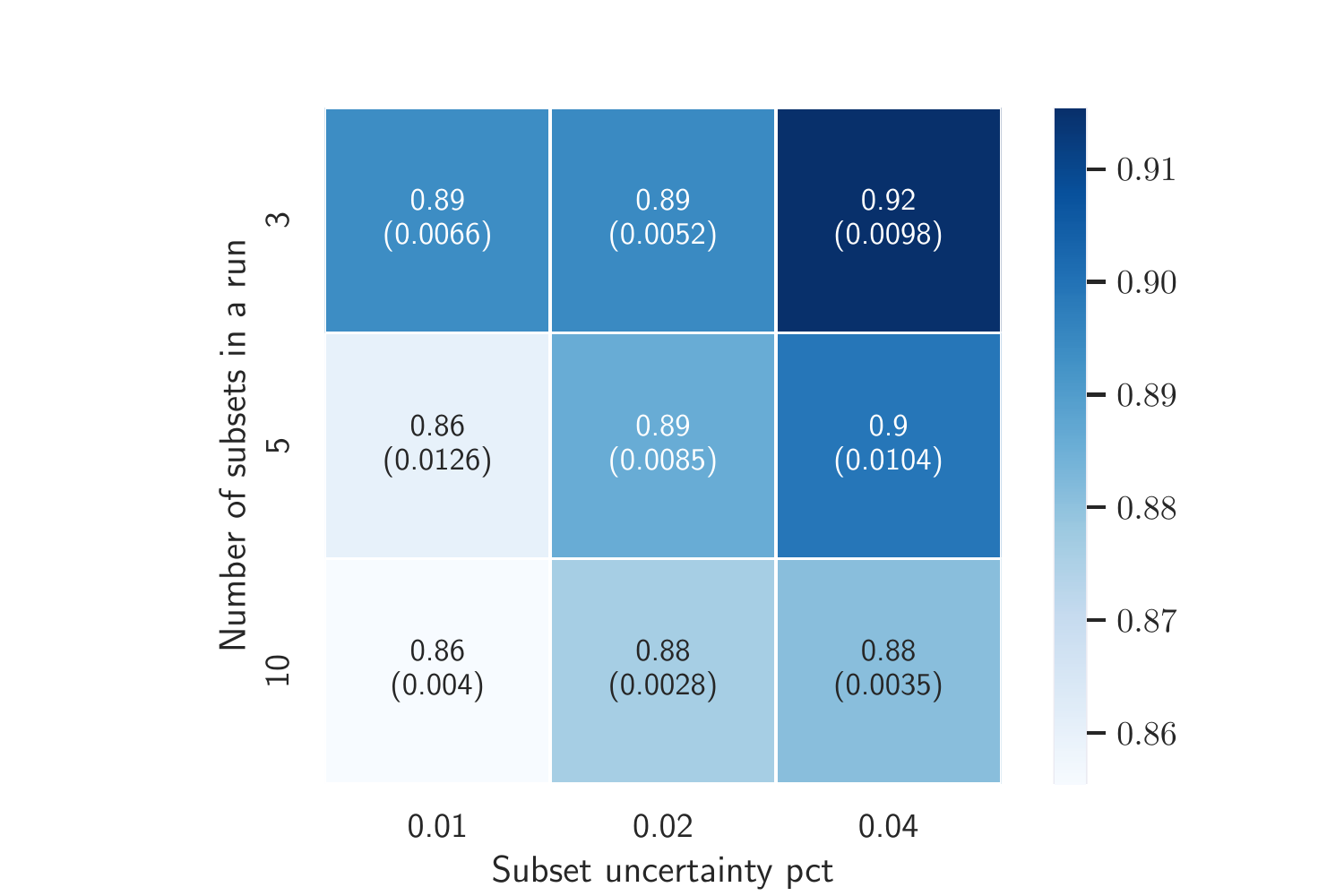} 
        \caption*{(b)}
    \end{minipage}
    \begin{minipage}{0.5\textwidth}
        \centering
        \includegraphics[width=\textwidth]{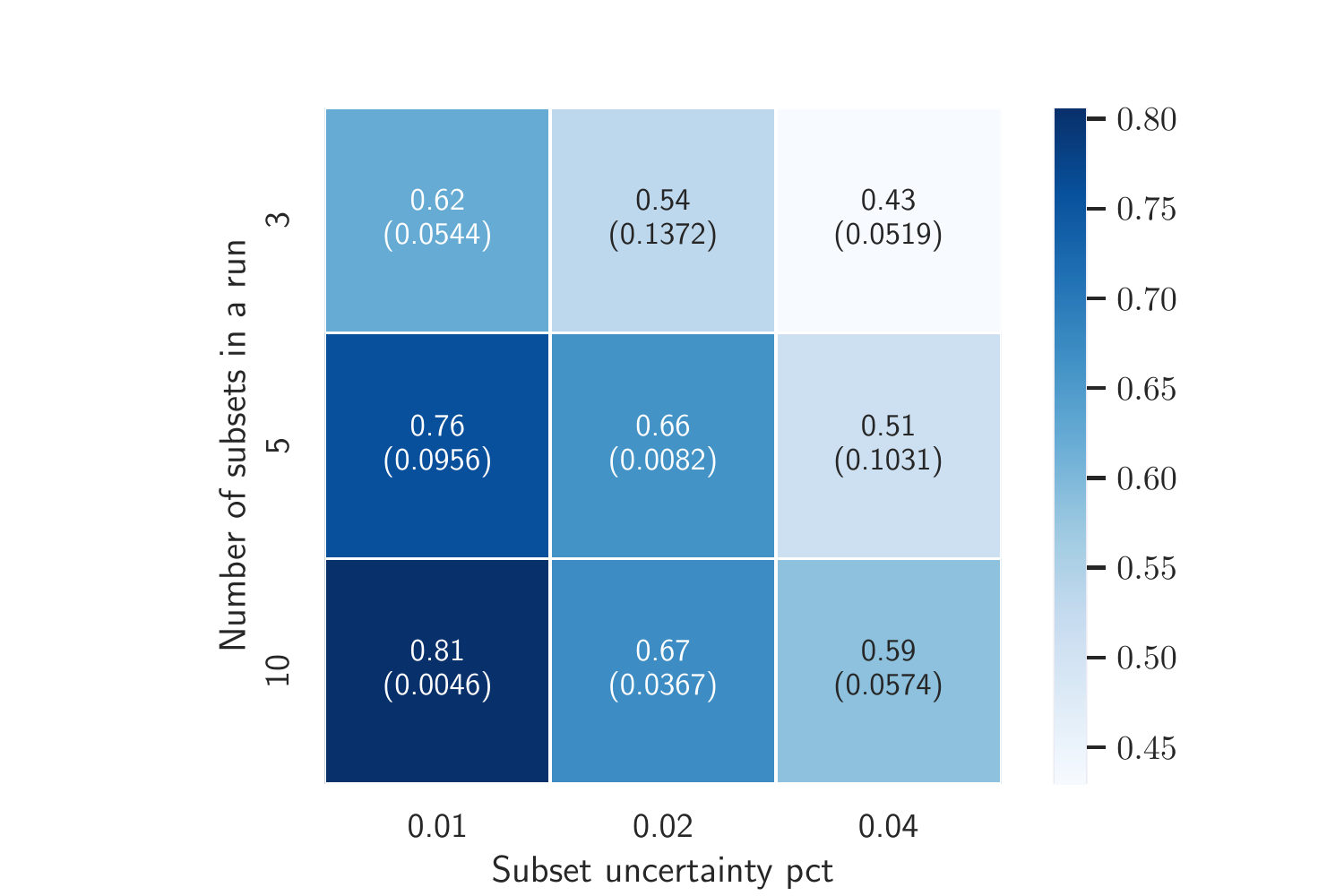} 
        \caption*{(c)}
    \end{minipage}
    \caption{Average over 3 independent runs for each of the criteria (Neuron Coverage (a), Similarity (b), Kill Ratio (c)) given $100$ evaluations when using the calibration set. Numbers in between parenthesis are the standard deviation.}
\end{figure}

\begin{figure}
    \centering
    \begin{minipage}{0.5\textwidth}
        \centering
        \includegraphics[width=\textwidth]{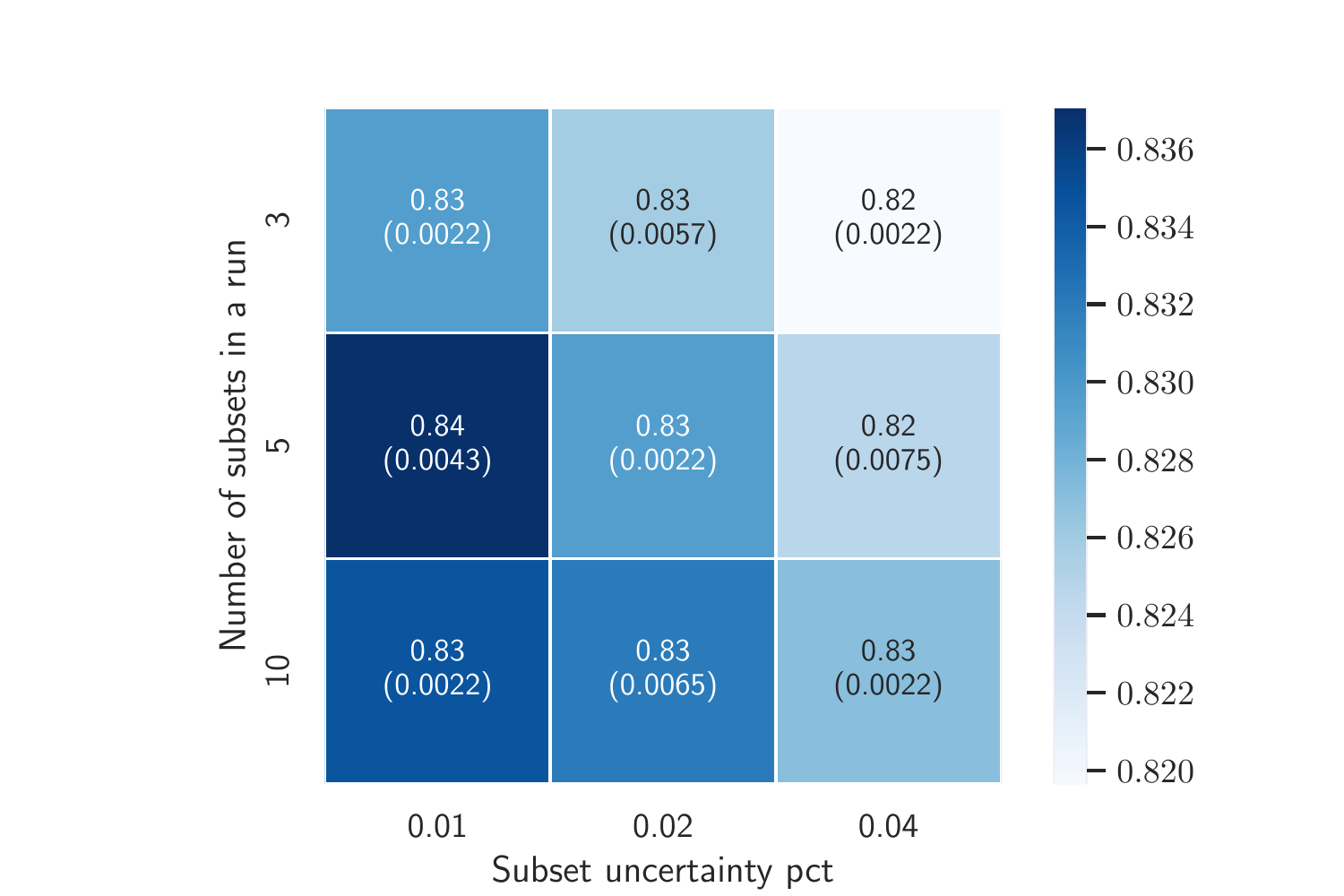} 
        \caption*{(a)}
    \end{minipage}\hfill
    \begin{minipage}{0.5\textwidth}
        \centering
        \includegraphics[width=\textwidth]{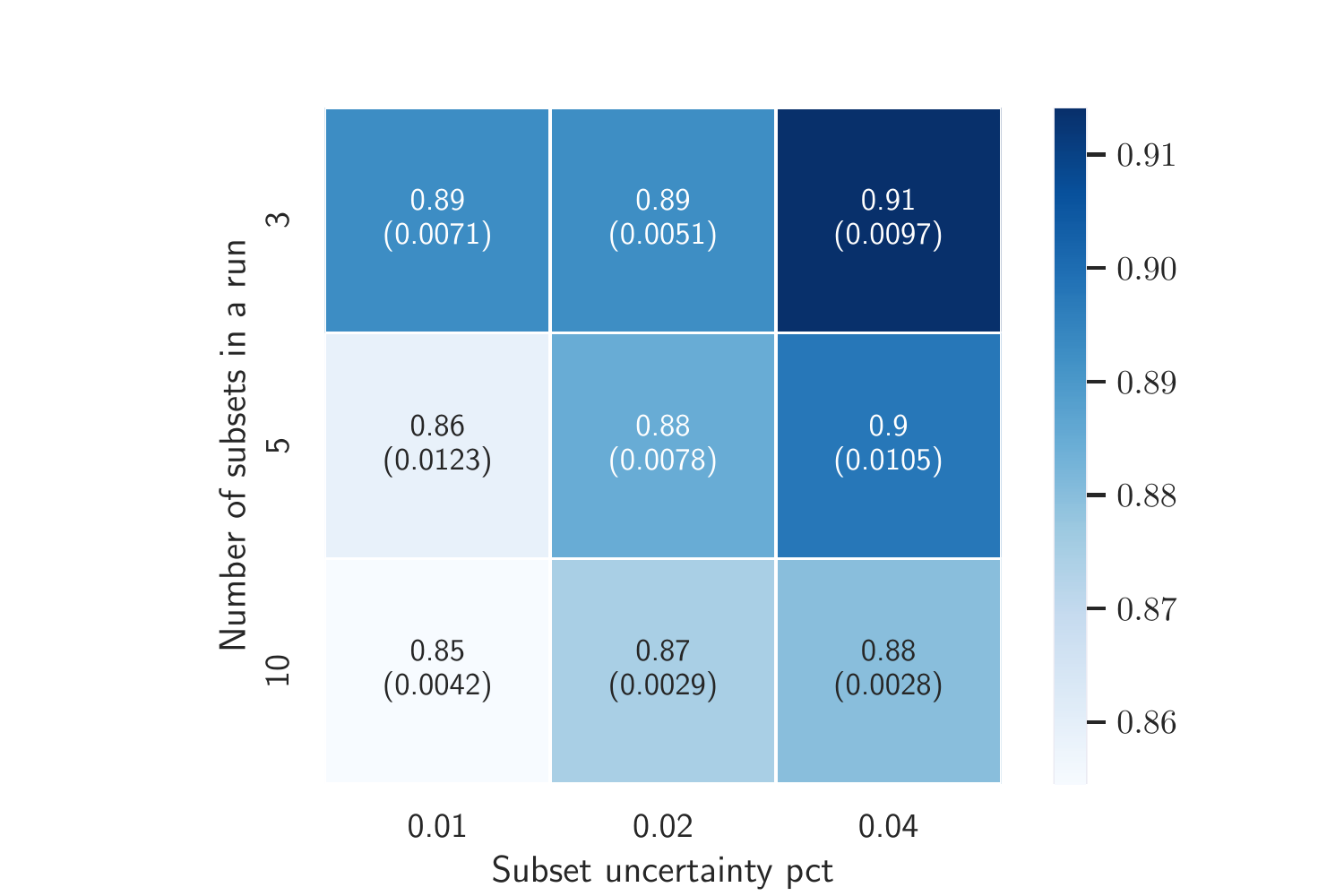} 
        \caption*{(b)}
    \end{minipage}
    \begin{minipage}{0.5\textwidth}
        \centering
        \includegraphics[width=\textwidth]{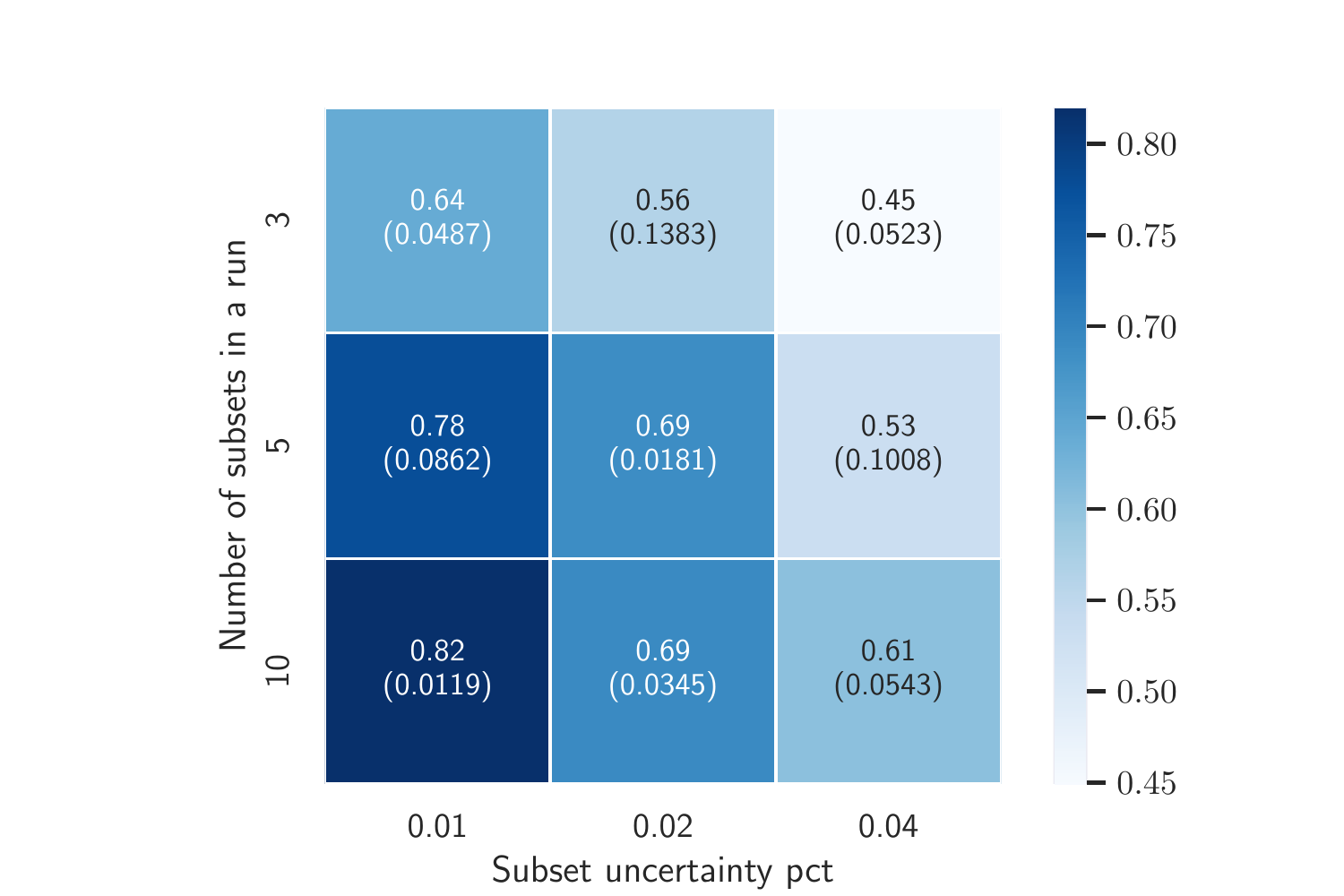} 
        \caption*{(c)}
    \end{minipage}
    \caption{Average over 3 independent runs for each of the criteria (Neuron Coverage (a), Similarity (b), Kill Ratio (c)) given $100$ evaluations when using the test set. Numbers in between parenthesis are the standard deviation.}
\end{figure}

\begin{figure}
    \centering
    \begin{minipage}{0.5\textwidth}
        \centering
        \includegraphics[width=\textwidth]{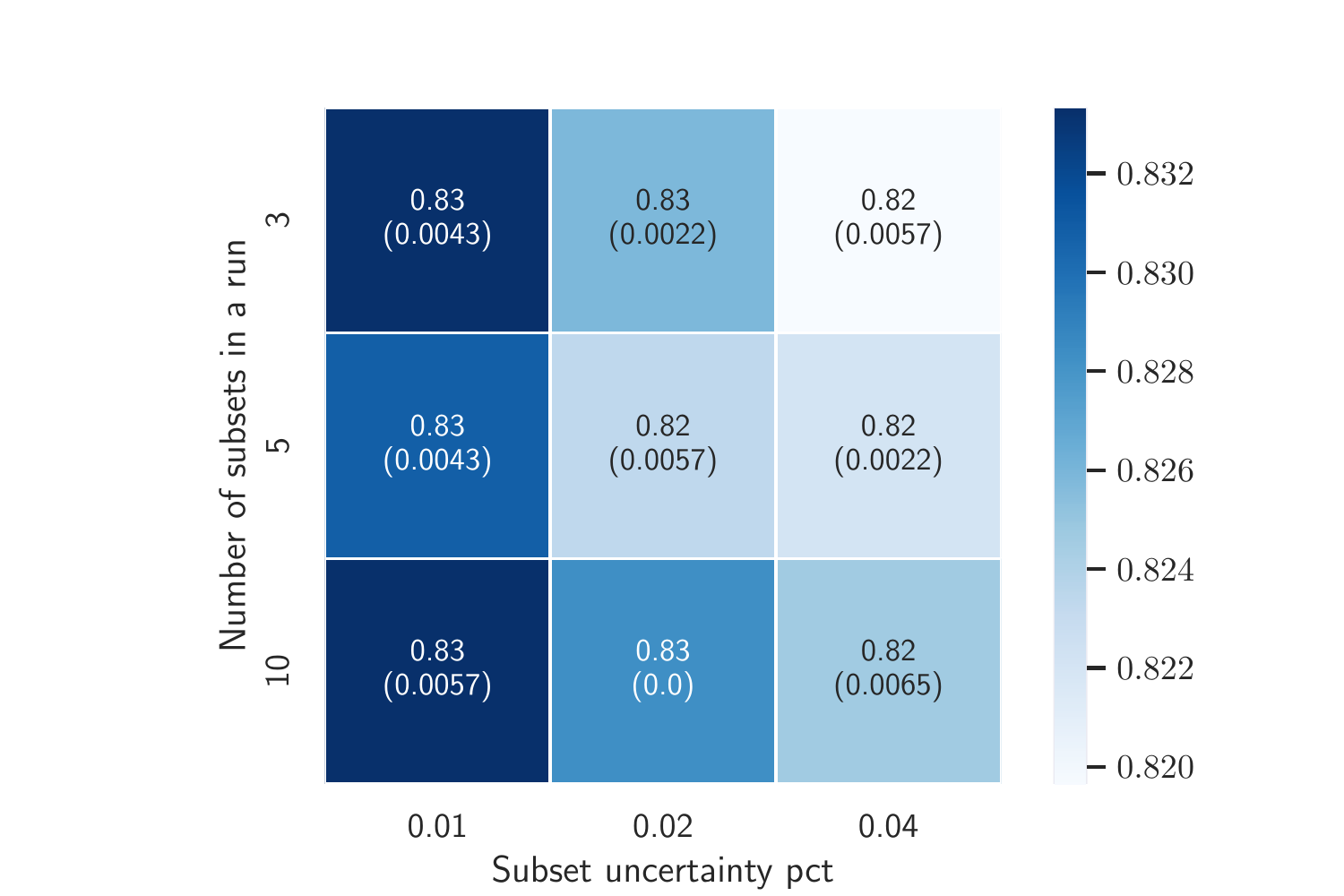} 
        \caption*{(a)}
    \end{minipage}\hfill
    \begin{minipage}{0.5\textwidth}
        \centering
        \includegraphics[width=\textwidth]{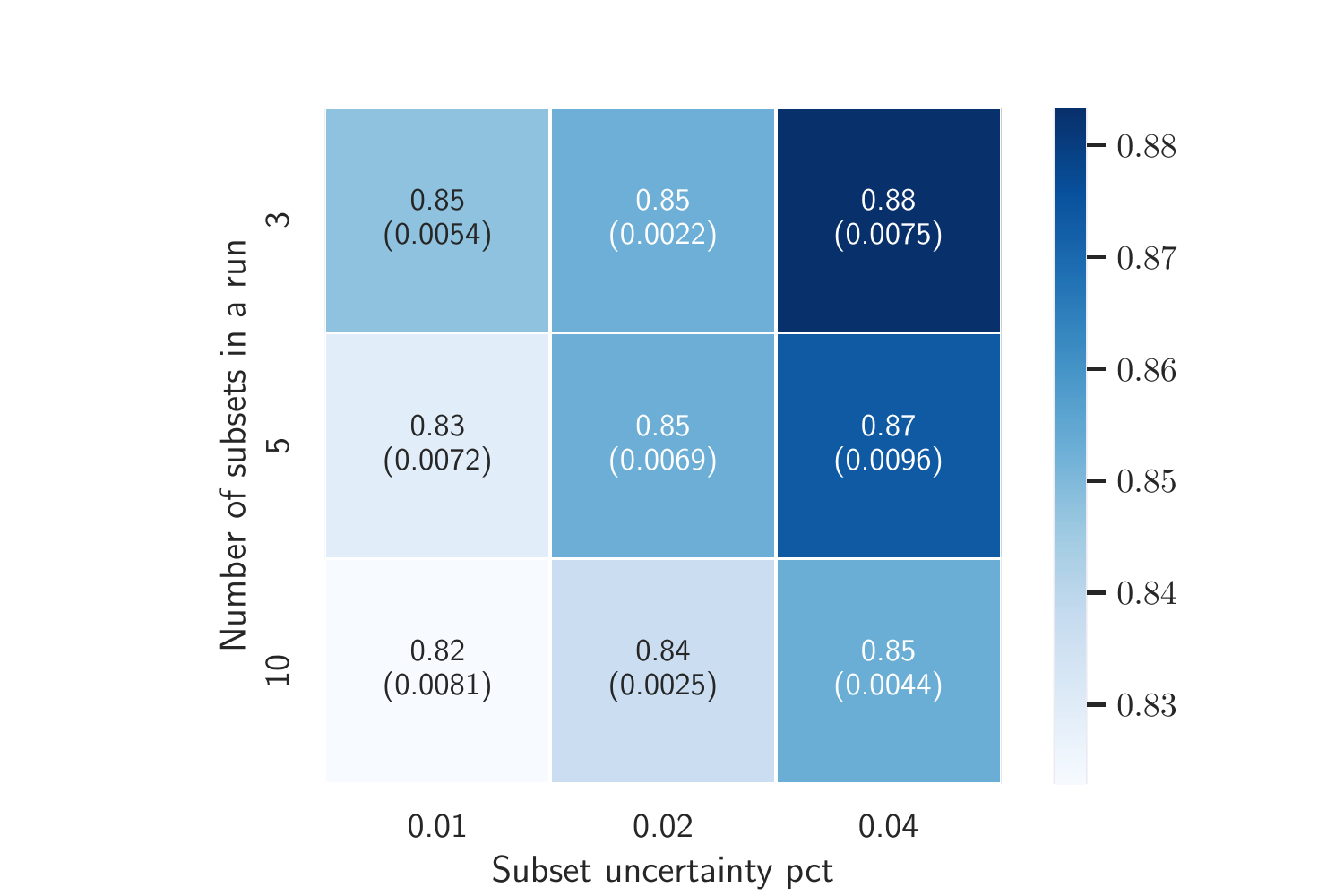} 
        \caption*{(b)}
    \end{minipage}
    \begin{minipage}{0.5\textwidth}
        \centering
        \includegraphics[width=\textwidth]{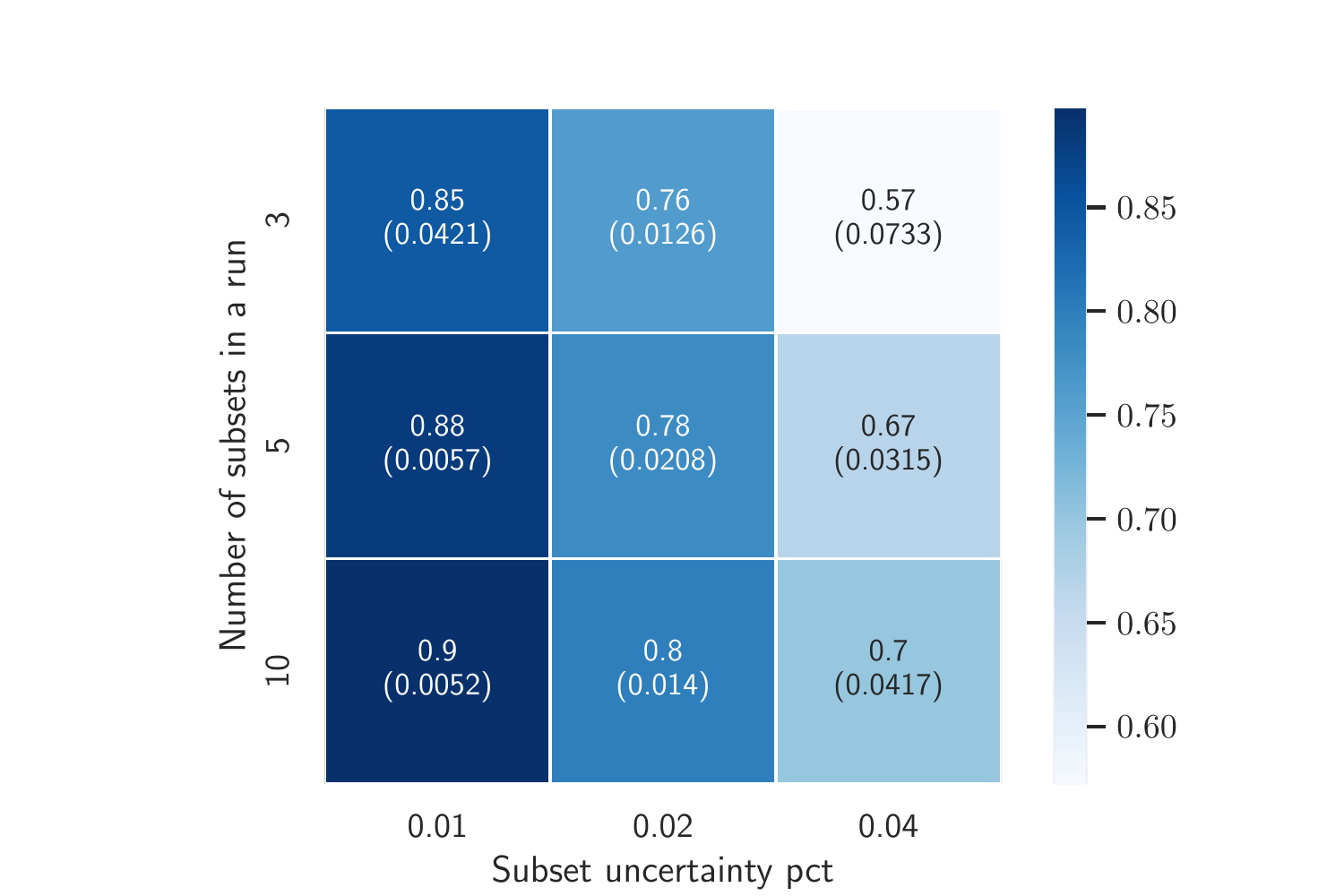} 
        \caption*{(c)}
    \end{minipage}
    \caption{Average over 3 independent runs for each of the criteria (Neuron Coverage (a), Similarity (b), Kill Ratio (c)) given $400$ evaluations when using the calibration set. Numbers in between parenthesis are the standard deviation.}
\end{figure}

\begin{figure}
    \centering
    \begin{minipage}{0.5\textwidth}
        \centering
        \includegraphics[width=\textwidth]{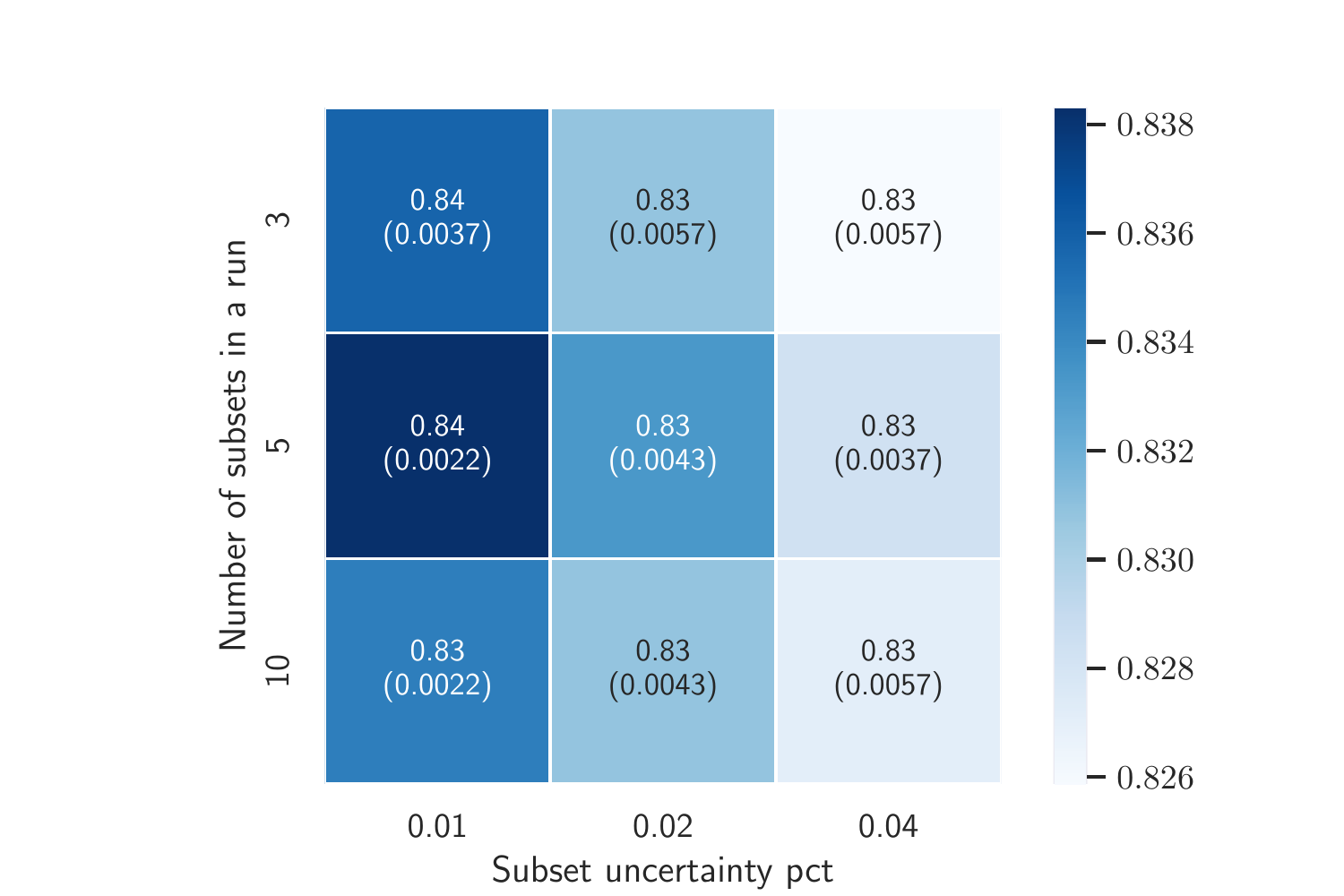} 
        \caption*{(a)}
    \end{minipage}\hfill
    \begin{minipage}{0.5\textwidth}
        \centering
        \includegraphics[width=\textwidth]{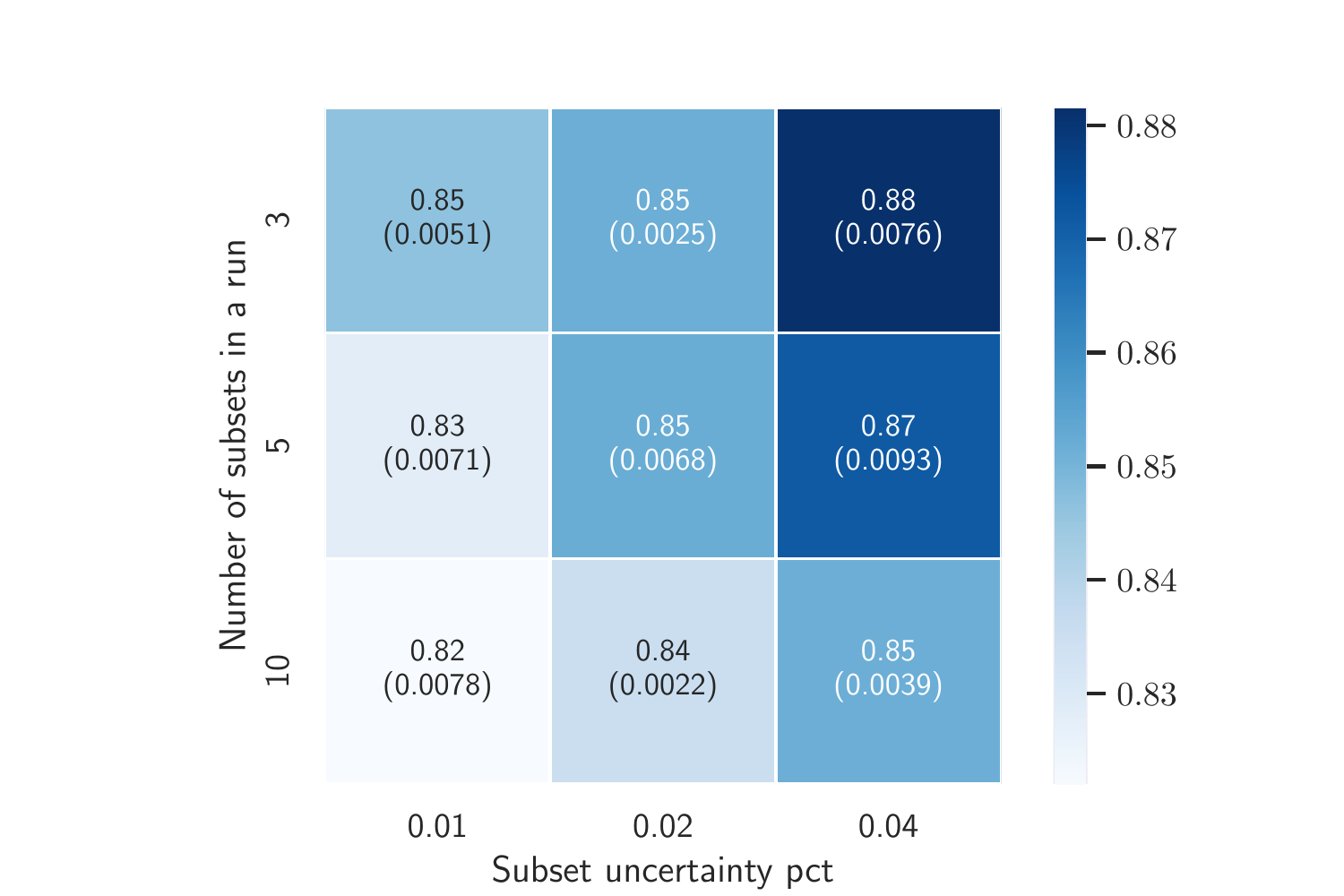} 
        \caption*{(b)}
    \end{minipage}
    \begin{minipage}{0.5\textwidth}
        \centering
        \includegraphics[width=\textwidth]{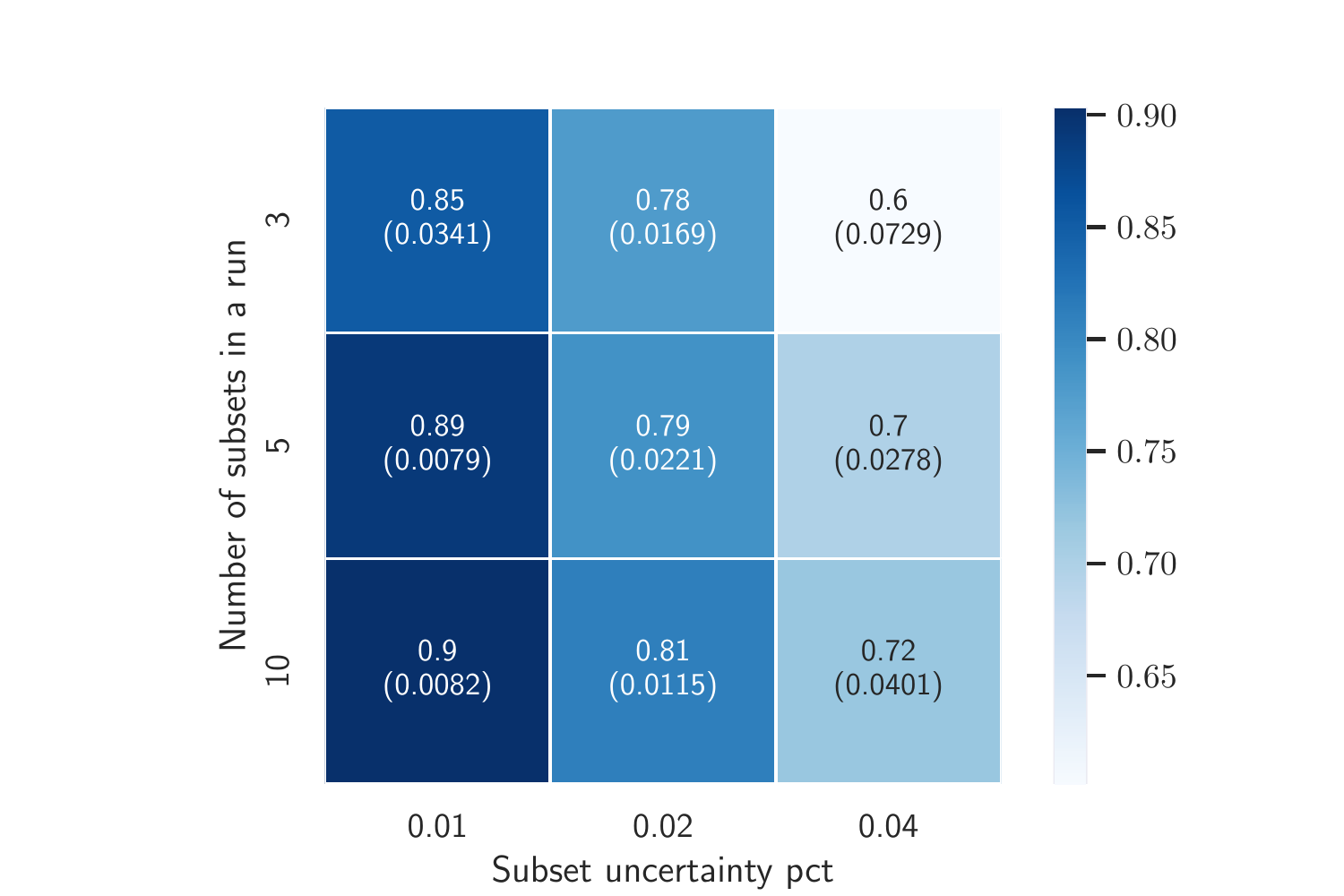} 
        \caption*{(c)}
    \end{minipage}
    \caption{Average over 3 independent runs for each of the criteria (Neuron Coverage (a), Similarity (b), Kill Ratio (c)) given $400$ evaluations when using the test set. Numbers in between parenthesis are the standard deviation.}
\end{figure}

\end{document}